\documentclass[journal]{IEEEtran}
\usepackage{lineno,hyperref}
\hypersetup{hidelinks,
	colorlinks=true,
	allcolors=black,
	pdfstartview=Fit,
	breaklinks=true}
\usepackage{bbm}
\usepackage{bm}
\usepackage{amsfonts}
\usepackage{amsmath}
\usepackage{mathrsfs}
\usepackage{amssymb}
\usepackage{enumerate}
\usepackage{graphicx}
\usepackage{subfigure}
\usepackage{booktabs}
\usepackage{graphicx}
\usepackage{setspace}
\usepackage{algorithm}
\usepackage{algorithmic}
\usepackage{setspace}
\usepackage{multirow}
\usepackage{color}
\usepackage{cite}
\usepackage{array}
\usepackage{times}
\usepackage{adjustbox} 
\usepackage{mathptmx}
\usepackage{dsfont}
\usepackage{booktabs} 
\usepackage{tabularx} 
\usepackage{diagbox}

\usepackage{makecell}  
\ifCLASSINFOpdf
\else
\fi
\begin{document}
	\title{
		Dual-task Mutual Reinforcing Embedded Joint Video Paragraph Retrieval and Grounding
	}
	\author{Mengzhao Wang, Huafeng Li,  Yafei Zhang,  Jinxing Li, Minghong Xie, Dapeng Tao \IEEEmembership{}  
		\thanks{This work was supported in part by the National Natural Science Foundation of China under Grant 62276120, and the Yunnan Fundamental Research Projects (202301AV070004, 202401AS070106).}
		\thanks{ M. Wang, H. Li, Y. Zhang, and M. Xie are with the Faculty of Information Engineering and Automation, Kunming University of Science and Technology, Kunming 650500, China.(E-mail: mengzhaowangg@163.com (M. Wang); lhfchina99@kust.edu.cn (H. Li)); zyfeimail@163.com (Y. Zhang); minghongxie@163.com (M. Xie)}
		\thanks{J. Li is with the School of Computer Science and Technology, Harbin Institute of Technology Shenzhen, 518055, China.}
		\thanks{D. Tao is with FIST LAB, School of Information Science and Engineering, Yunnan University, Kunming 650091, China.}
\thanks{Manuscript received xxxx;}}
\markboth{Journal of \LaTeX\ Class Files}%
{Shell \MakeLowercase{\textit{et al.}}}
\maketitle
\begin{abstract}
Video Paragraph Grounding (VPG) aims to precisely locate the most appropriate moments within a video that are relevant to a given textual paragraph query. However, existing methods typically rely on large-scale annotated temporal labels and assume that the correspondence between videos and paragraphs is known. This is impractical in real-world applications, as constructing temporal labels requires significant labor costs, and the correspondence is often unknown. To address this issue, we propose a Dual-task Mutual Reinforcing Embedded Joint Video Paragraph Retrieval and Grounding method (DMR-JRG). In this method,  retrieval and grounding tasks are mutually reinforced rather than being treated as separate issues. DMR-JRG mainly consists of two branches: a retrieval branch and a grounding branch. The retrieval branch uses inter-video contrastive learning to roughly align the global features of paragraphs and videos, reducing modality differences and constructing a coarse-grained feature space to break free from the need for correspondence between paragraphs and videos. Additionally, this coarse-grained feature space further facilitates the grounding branch in extracting fine-grained contextual representations. In the grounding branch, we achieve precise cross-modal matching and grounding by exploring the consistency between local, global, and temporal dimensions of video segments and textual paragraphs. By synergizing these dimensions, we construct a fine-grained feature space for video and textual features, greatly reducing the need for large-scale annotated temporal labels. Meanwhile, we design a grounding reinforcement retrieval module (GRRM) that brings the coarse-grained feature space of the retrieval branch closer to the fine-grained feature space of the grounding branch, thereby reinforcing retrieval branch through grounding branch, and finally achieving mutual reinforcement between tasks. Extensive experiments on three challenging datasets demonstrate the effectiveness of our proposed method. The code is available at \href{https://github.com/X7J92/DMR-JRG}{\textcolor{blue}{https://github.com/X7J92/DMR-JRG}}.

\end{abstract}
\IEEEpeerreviewmaketitle
\section{Introduction}

\begin{figure}[t!]
\centering

\includegraphics[width=0.95\linewidth]{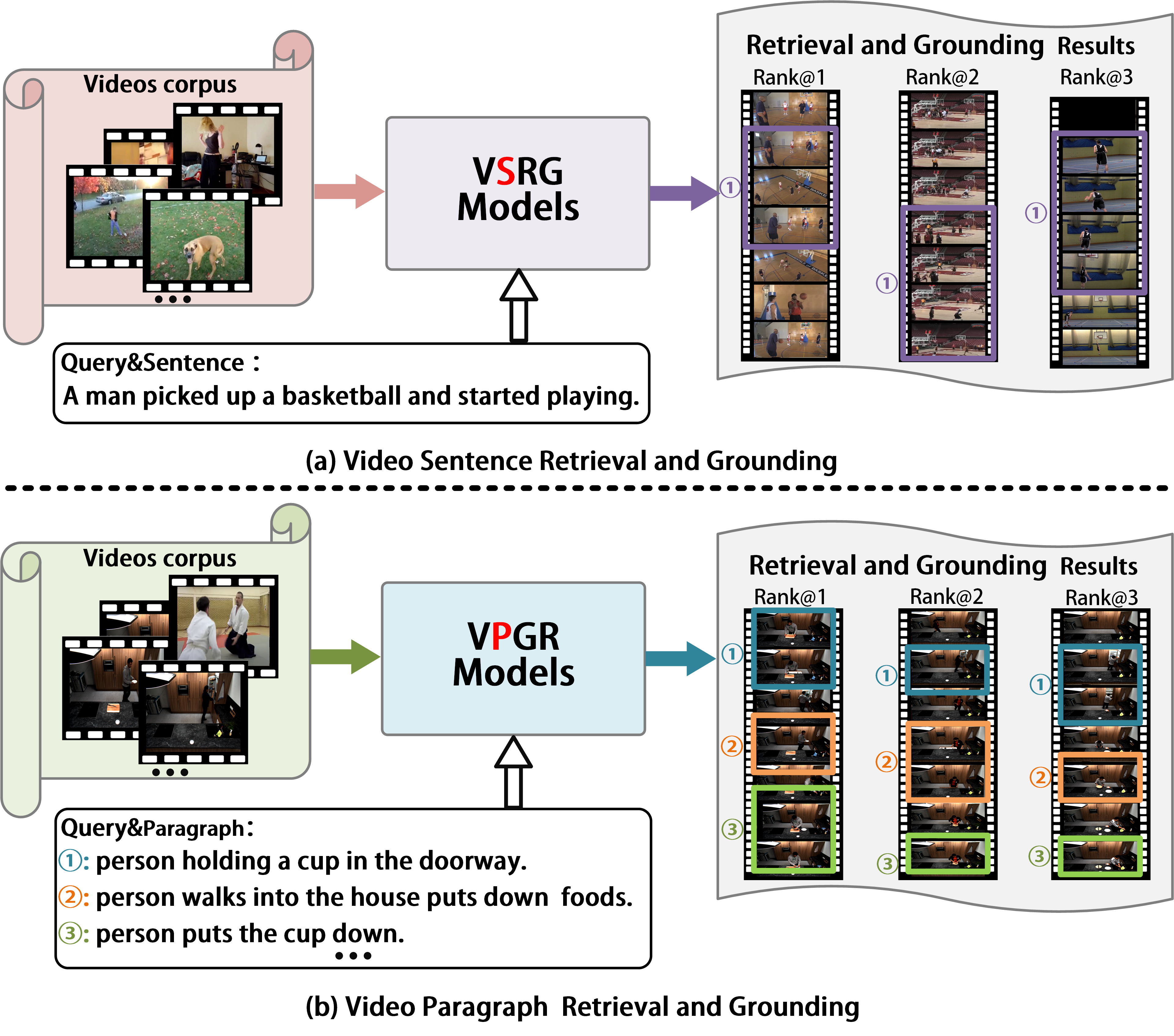}
\caption{An illustrative example of Video Sentence Retrieval and Grounding (VSRG) and Video Paragraph Retrieval and Grounding (VPRG). (a) VSRG aims to retrieve the corresponding video from a video corpus along with its specific moments using a single sentence as a query. (b) VPRG aims to retrieve the relevant video from a video corpus through a  paragraph query and to locate the exact moments of each sentence in the paragraph within the video.}

\label{Fig01}
\end{figure}

Video grounding aims to precisely locate the most relevant video segments within untrimmed video streams, guided by specific textual descriptions. This technology is widely applied in areas such as 
video anomaly detection\cite{sultani2018real,wu2024toward}, target tracking\cite{liu2023robust,wang2022yolov3,liu2020deepmtt}, object detection\cite{8691693,9905640,carion2020end,sun2021sparse,cai2018cascade,leng2024recent,10252044} and can also be used as an alternative to person re-identification\cite{li2021triple,li2023logical,chenyiwen2020,li2023intermediary}. Traditionally, video grounding research has primarily focused on Video Sentence Grounding (VSG) \cite{cui2022video,paul2021text,9953557,10321681,10214636,9978670,10605104,9882521}, which aligns single sentences with corresponding video segments. However, in real-world applications, a more powerful and effective approach is often required, which entails utilizing text paragraphs to accurately pinpoint and retrieve pertinent video segments. This need has led to the emergence of Video Paragraph Grounding (VPG) \cite{tan2024siamese,bao2021dense,jiang2022semi,jiang2022gtlr,tan2023hierarchical,shi2024end}. When compared to VSG, VPG presents noteworthy advantages thanks to the abundant contextual information conveyed by text paragraphs. This enhanced context allows for a more nuanced capture and accurate mapping of intricate video content, thereby attaining higher-quality matching and precise grounding in real-world applications.

DepNet\cite{bao2021dense} was the first method for tackling the VPG task, based on a supervised learning paradigm. The method explored the contextual relationships between sentences within a paragraph, aiming to identify relevant video segments or candidate moments for each sentence. To further address the challenges of contextual modeling with complex paragraph queries, GTLR \cite{jiang2022gtlr} introduced a novel framework for VPG. This framework was the first to jointly reason over video and paragraph content using a multimodal graph encoder. Nonetheless, achieving a comprehensive understanding of the relationship between video content and the overall meaning conveyed by an entire paragraph remains challenging. In addition, these methods are typically developed within a supervised learning paradigm and rely on large-scale annotated temporal labels, which results in high manual costs. These issues indicate that further research is needed in this field to overcome the limitations of existing methods.

Recently, The PRVG method \cite{shi2024end} attempts to apply the Transformer to VPG, aiming to unravel the intricate connections between video content and the entirety of a paragraph. In this approach, textual features are extracted from the whole paragraph to serve as queries, while video features are used as keys and values. Mirroring the Transformer's architecture, it employs end-to-end parallel decoding to precisely pinpoint the temporal boundaries of the content referenced by each sentence in the video. However, a notable limitation of this method is its insufficient consideration of the sequential impact of sentence order. This may lead to decreased accuracy in aligning video segments with the corresponding sentences in a paragraph.  Meanwhile, existing VPG methods often assume known correspondence between videos and paragraphs in training and testing. However, in real-world applications, such correspondences are usually unknown, requiring the retrieval of relevant videos from a video corpus before performing the VPG task. This brings new challenges to VPG task.

Additionally, JSG\cite{chen2023joint} introduced a new task: Video Sentence Retrieval and Grounding (VSRG). VSRG aims to use a single sentence as a query to retrieve the corresponding video from a video corpus and precisely locate the relevant moment (as shown in Fig. \ref{Fig01}(a)). Compared to previous VSG methods, VSRG adds retrieval capabilities, aligning more closely with real-world application needs. Although JSG performs well in VSRG tasks, it is specifically designed for single sentence queries and does not take into account the complexities of paragraph queries and the contextual relationships between sentences within a paragraph. This leads to JSG's inability to accurately retrieve and locate video segments related to the entire paragraph content, making solving Video Paragraph Retrieval and Grounding task a significant challenge. To  address the these previously mentioned challenges, it is essential to develop a method that not only has video paragraph retrieval capabilities but also provides a comprehensive and nuanced understanding of the interactions between video content and the entire paragraph during video paragraph grounding, while considering the impact of sentence sequencing. Additionally, exploring weakly supervised learning techniques during the grounding stage is crucial for reducing annotation costs and enhancing the model's generalizability and practical applicability.

\begin{figure}[t!]
\centering
\includegraphics[width=0.80\linewidth]{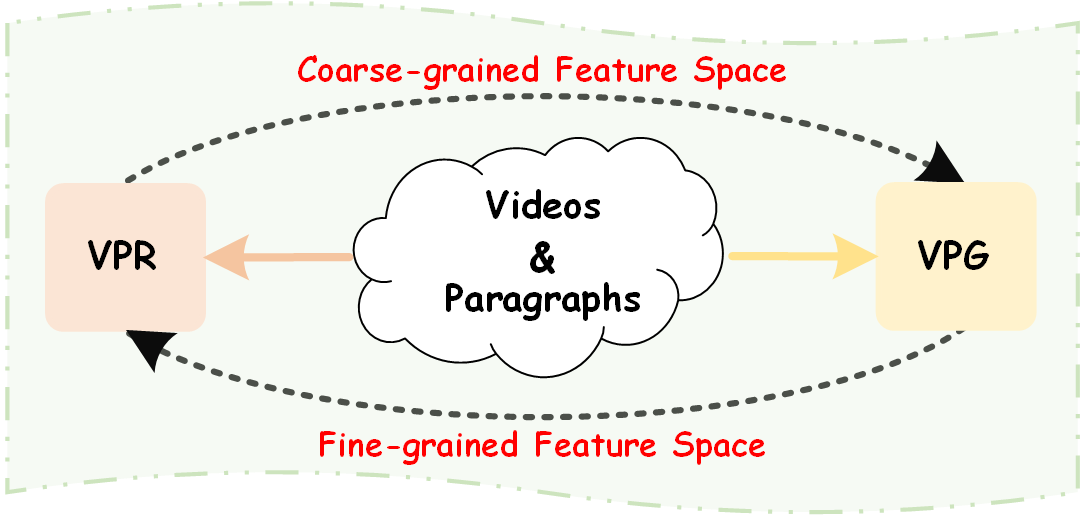}
\caption{Details of the dual-task mutual reinforcing framework. VPR stands for Video Paragraph Retrieval, and VPG stands for Video Paragraph Grounding.}

\label{Fig02}
\end{figure}

In this paper, we define a new task called Video Paragraph Retrieval and Grounding (VPRG) as shown in Fig. \ref{Fig01}(b). VPRG aims to retrieve the relevant video from a video corpus through a textual paragraph query and to locate the exact moments of each sentence in the paragraph within the video.  To address the VPRG task, we proposes Dual-task Mutual Reinforcing Embedded Joint Video Paragraph Retrieval and Grounding method (DMR-JRG). The DMR-JRG framework is shown in Fig. \ref{Fig02}. Specifically, this method is mainly divided into two branches: a retrieval branch and a grounding branch. In the retrieval branch, we use inter-video contrastive learning to roughly align paragraph and video features, thus constructing a coarse-grained feature space. This  space lays the foundation for fine-grained feature alignment in the grounding branch, reinforcing more precise cross-modal matching and grounding. In the grounding branch, our approach frees itself from the constraints of traditional supervised learning methods, which typically rely on large amounts of temporal labels. Instead, by deeply exploring the multi-dimensional congruity between videos and text paragraphs across local, global, and temporal dimensions, it achieves precise cross-modal grounding. Technically speaking, our method integrates the complementary nature of visual and textual features across local, global, and temporal dimensions, effectively bridging the modality gap to achieve feature consistency. Within the local dimension, we meticulously establish correspondences between individual sentences in textual paragraphs and their corresponding video segments, providing a solid foundation for subsequent cross-modal matching. In the global dimension, we introduce a special token, known as the class token, specifically designed to integrate and represent overarching information from both textual and visual modalities at a higher level. This class token plays a pivotal role in encapsulating global context, ensuring a semantically robust alignment between the two modalities. Finally, in the temporal dimension, we delve into the sequential patterns hidden within sentences, further refining the matching between specific sentences and their related video features for notably enhanced VPG accuracy. By synergizing these dimensions, we construct a fine-grained feature space for video and textual features, greatly reducing the need for fully annotated temporal labels.

Additionally, to reinforce the retrieval branch through the grounding branch, we design a Grounding Reinforcement Retrieval Module (GRRM). Specifically, we obtain cosine similarity scores between videos and paragraphs from the grounding branch and use these scores as pseudo-labels, along with the MSE loss function, to constrain the cosine similarity scores of the retrieval branch. GRRM allows the coarse-grained feature  space of the retrieval branch to more closely align with the fine-grained feature space of the grounding branch, thereby effectively reinforcing the retrieval branch through grounding branch.

In summary, the significant contributions of this research are outlined below:\vspace{-0.0mm}
\begin{itemize}
\item   We define a new task called Video Paragraph Retrieval and Grounding (VPRG) and develop a Dual-task Mutual Reinforcing Embedded Joint Video Paragraph Retrieval and Grounding  (DMP-JRG) method as the first attempt of tackling this task. Particularly, DMP-JRG adopts a novel scheme of mutual reinforcing of retrieval and grounding, which effectively solves the VPRG task even in the absence of annotated temporal labels and when the correspondence between videos and paragraphs is unknown.

\item  In the grounding branch, we present a novel approach that seamlessly integrates three crucial techniques. First, local dimension congruity links sentences to their corresponding video segments. Second, global dimension alignment uses category labels to align text and video data. Finally, temporal dimension synchronization refines matching results by taking sentence order into account. Together, these advancements significantly boost the accuracy and reliability of VPG in a weakly supervised setting.

\item Experiments on the large-scale datasets ActivityNet Captions, Charades-STA, and TaCoS demonstrate that our proposed method significantly outperforms several latest VSRG methods.
\end{itemize}

\section{Related Work}
\subsection{Video Sentence Grounding}
VSG focuses on matching individual sentences with the content in the video. The primary challenge confronted by these methods is how to precisely comprehend and align semantic information from visual elements in the video with that of the query sentences. Typically, VSG approaches can be categorized into two types depending on whether the proposal generation module is employed: proposal-based methods\cite{zhangSongyang2020,zhang2021multi,wang2021structured,zheng2023progressive,gao2021fast,10023968} and non-proposal-based\cite{10239444,10449438,10233190,mun2020local,li2021proposal,xu2022hisa,cui2022video,paul2021text} methods. Proposal-based methods follow a two-stage procedure, initially generating candidate video segments and then jointly modeling those segments alongside sentence queries to identify the segment that best matches the query. In contrast, non-proposal-based methods aim to directly encode visual and textual information through feature extraction networks, enhancing comprehension by aggregating contextual information. These methods commonly leverage multimodal attention mechanisms for cross-modal inference and feed the resulting multimodal features into regressors to predict the relevant moments.

In addition, there are also some weakly supervised methods in the field of VSG. These methods require only pairs of videos and queries during the training process, eliminating the need for precise start/end time annotations. Generally speaking, existing weakly supervised VSG methods can be further categorized into two types: those based on multi-instance learning \cite{wang2022siamese,wang2021weakly,da2021asynce,mithun2019weakly} and those based on reconstruction \cite{10374139,lin2020weakly,lv2023counterfactual,zheng2022weakly1,zheng2022weakly2}. Among these, TGA \cite{mithun2019weakly} stands out as the pioneer in applying multi-instance learning frameworks to address weakly supervised problems. It learns global video representations for specific texts by encoding videos, extracting query features, and utilizing text-guided attention mechanisms. Subsequently, it maps both visual and textual features into a shared space, achieving alignment of visual and textual elements. 
In contrast, methods based on reconstruction tackle weak supervision indirectly. They first generate candidate segments that match the query based on the input video and query, and then use these segments to reconstruct the query, thereby achieving grounding. A representative example of this approach is SCN \cite{lin2020weakly} which retrieves a set of candidate segments from the video to reconstruct the masked query, selecting the most relevant segment as input. The reward calculated based on reconstruction loss is further used to optimize the generation process of these segments. However, the majority of these methods are limited to addressing the grounding of a single event and struggle to simultaneously locate multiple events.

Additionally, JSG\cite{chen2023joint} first introduced the VSRG task. VSRG aims to use a single sentence as a query to retrieve the corresponding video and precisely locate the relevant moment. Specifically, the JSG method addresses VSRG through a glance-to-gaze process and a synergistic contrastive learning strategy. Although JSG performs well in VSRG task, it cannot solve the VPRG task.

\begin{figure*}[t!]
\centering
\includegraphics[width=\textwidth]{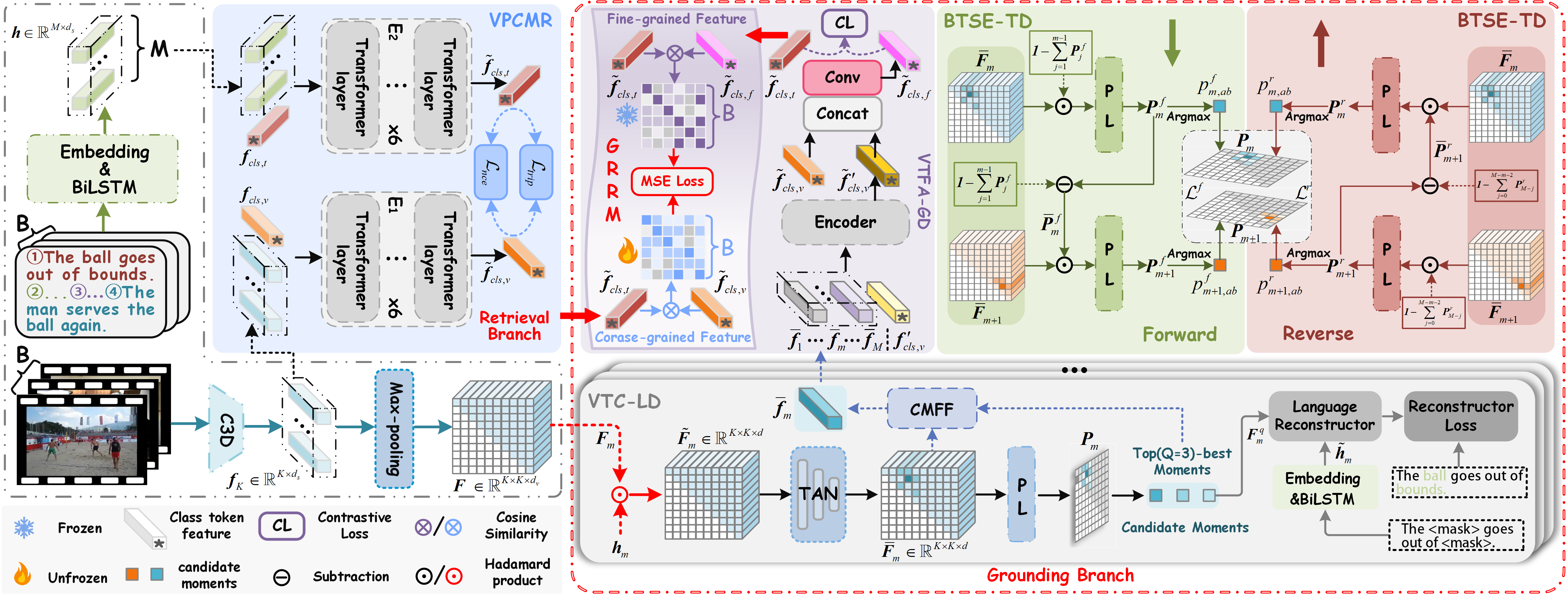}

\caption{Overview of the proposed method. It comprises there core parts: Firstly, Feature Extraction, which includes video feature extraction and text feature extraction. Secondly, Retrieval Branch, consists of the Video-Paragraph Cross Modal Retrieval (VPCMR) component. Lastly, Grounding Branch, which includes three components: Visual-Textual Consistency on local dimension (VTC-LD), Visual-Textual Feature Alignment on global dimension (VTFA-GD), and Bidirectional Temporal Synchronization of Events on temporal dimension (BTSE-TD). By combining the strengths of these three parts, we  mitigate the differences between visual and textual modalities, ensuring that the text accurately corresponds to events in the video. This effectively achieves video paragraph retrieval and grounding.}
\vspace{-14pt} 
\label{Fig03}
\end{figure*}

\subsection{Video Paragraph Grounding}
Compared to the VSG, the VPG poses a greater challenge. It requires precisely grounding all relevant moments in untrimmed videos based on paragraphs describing multiple events. Presently, VPG methodologies can broadly be categorized into fully supervised\cite{bao2021dense,jiang2022semi,jiang2022gtlr,tan2023hierarchical,shi2024end} and weakly supervised approaches\cite{tan2024siamese}. Under fully supervised settings, the training sentences align closely with their respective labels, thereby offering clear guidance for the model training. DepNet \cite{bao2021dense}, as the pioneering approach in the VPG field, adeptly incorporates both local and global information within paragraphs to enhance the precision of grounding multiple events. Specifically, it generates both visual and semantically-driven temporal proposals for individual paragraph events and unifies them using second-order attention. The model then selects the most pertinent subset of features for each proposal and facilitates the integration of global and local proposal features through a soft attention mechanism. This methodology ensures that each event proposal is based on its own description while also leveraging the temporal context of other paragraph events, resulting in precise grounding within intricate videos. 
Similarly, the HSCNet \cite{tan2023hierarchical} aims to explore multi-level visual-textual correspondence by learning hierarchical local-global semantic alignment to address the video paragraph grounding task.

However, a significant drawback of fully supervised methods is their heavy reliance on extensive annotated datasets, which restricts the broad applicability of the model. To mitigate this challenge, SiamGTR \cite{tan2024siamese} propose a novel siamese learning framework to
jointly learn the cross-modal feature alignment and temporal coordinate regression without ground-truth supervision. Nonetheless, despite achieving notable progress, this method solely considers the relationship between text paragraphs and videos from a singular perspective, neglecting the complementary nature of multi-dimensional features. Meanwhile, existing VPG methods\cite{tan2024siamese,bao2021dense,jiang2022semi,jiang2022gtlr,tan2023hierarchical,shi2024end} assume that the correspondence between videos and paragraphs is known during both training and testing. However, in real-world applications, such correspondences are often unknown, requiring the retrieval of relevant videos from a video corpus before performing the VPG task. To address these limitations, we propose a novel Dual-task Mutual Reinforcing Embedded Joint Video Paragraph Retrieval and Grounding framework, named DMR-JRG. Our framework requires only the knowledge of paragraph-video correspondence during training, eliminating the need for any additional temporal labels. By leveraging the mutual reinforcement between retrieval and grounding tasks, we advance the state-of-the-art in VPRG.

\section{The Proposed Method}
\subsection{Problem Definition}

Here we first give a specific definition for the VPRG task. Assume that we have a video corpus containing a set of $N$ untrimmed long videos, denoted as $ \bm{V} = \left\{ {{V_n}} \right\}_{n = 1}^N$ where $ V_n= \left\{ {{x_t}} \right\}_{t = 1}^T$ and $x_t$ denotes the $t$-th frame and $T$ is the total number of frames. A text paragraph query is represented as $\bm{S}=\{s_m\}_{m=1}^{M}$ with $s_m$ representing the $m$-th sentence in this paragraph. The goal of our task is to retrieve from this video corpus: (1) a video $V^*$ that contains the relevant events to the paragraph query and (2) produce a set of time intervals $\bm {T}=\{(t_s, t_e)_m\}_{m=1}^{M}$, where $t_s$ and $t_e$ respectively denote the start and end times of the event described by the $m$-th sentence in $V^*$. In this study, we consider a weakly supervised setting where the temporal boundary ($t_s$, $t_e$) between the sentences in $\bm{S}$ and the corresponding events in $V^*$ remain unknown during training. We only know the correspondence between $V^*$ and $\bm{S}$ during training but the correspondence is not available during testing. Our aim is to train a model that can accurately retrieve corresponding videos ($V^*$) based on text paragraph query ($\bm{S}$) from a video corpus and locate the start ($t_s$) and end ($t_e$) times of events described by each sentence in the paragraph within the video $V^*$. The lookup table for the main notations in this paper is shown in Table~\ref{tab1}.
\subsection{Overview}
The schematic representation of our proposed method is shown in Fig. \ref{Fig03}. This method consists of three parts: Feature Extraction, Retrieval Branch and Grounding Branch. Specifically, the Feature Extraction includes video feature extraction and text feature extraction, extracting the features of the input videos and paragraphs respectively. The Retrieval Branch is the Video-Paragraph Cross Modal Retrieval (VPCMR) module, which aims to achieve cross modal video-paragraph retrieval through the use of joint InfoNCE loss and triplet loss. The Grounding Branch includes three components: Visual-Textual Consistency on Local Dimension (VTC-LD), Visual-Textual Feature Alignment on Global Dimension (VTFA-GD), and Bidirectional Temporal Synchronization of Events on Temporal Dimension (BTSE-TD). These three components work together to explore the multi-dimensional congruity between video segments and textual paragraphs for precise cross-modal grounding.  

In our method, the retrieval branch constructs a unified coarse-grained feature space by aligning the global features of paragraphs and videos. This coarse-grained space provides a foundation for the grounding branch to extract finer-grained contextual representations. Furthermore, we design a Grounding Reinforcement Retrieval Module (GRRM) to align the coarse-grained feature space of the retrieval branch with the fine-grained feature space of the grounding branch. This design enables the retrieval branch to benefit from the grounding branch, thereby achieving mutual reinforcement between the two branches. Fig. \ref{Fig02} also clearly illustrates the mutual reinforcement process between the retrieval and grounding tasks.

\subsection{Feature Extraction}
\subsubsection{Text Feature Extraction} Given an input text paragraph $S$, we employ a Bidirectional Long Short-Term Memory (Bi-LSTM) \cite{hochreiter1997long} as the text encoder. Let the $m$-th sentence in the input be denoted as ${s_m} = \{ {s_{m,j}}\}_{j = 1}^J$, where $s_{m,j}$ represents the $j$-th word in the sentence, and $J$ is the total number of words in ${s_m}$. Each word ${s_{m,j}}$ in the sentence ${s_m}$ is embedded using Word2Vec  \cite{mikolov2013distributed} to obtain the word embedding ${{\bm{w}}_{m,j}} \in {\mathbb{R}^{{d_s}\times 1}}$, where ${d_s}$ is the dimensionality of each embedded word. The entire sentence's word embeddings $\{\bm{w}_{m,j}\}_{j = 1}^J$ are then fed into the bidirectional Bi-LSTM, and the final output ${{\bm{h}}_m} \in {\mathbb{R}^{1 \times {d_s}}}$ of the network is taken as the feature representation of the sentence ${s_m}$. To obtain cross-modal consistent features at the local level, following SCN\cite{lin2020weakly}, we randomly mask one-third of the words in the word embeddings of the $m$-th sentence, denoted as $\{{\bm{\hat w}}_{m,j}\}_{j = 1}^J$. Notably, the word embeddings for both the masked and complete sentences are encoded by the same Bi-LSTM. The word embeddings $\{{\bm{\hat w}}_{m,j}\}_{j = 1}^J$ are fed into the Bi-LSTM to obtain the feature representation ${\bm{\hat h}}_m \in {\mathbb{R}^{1 \times {d_s}}}$ of the $m$-th masked sentence.

\subsubsection{Video Feature Extraction}
For a video $V$, we utilize the 2D-TAN\cite{zhangSongyang2020} approach to construct candidate temporal segments, which represent potential moments of interest within the video.
Specifically, the input video $V$ is segmented into a sequence of non-overlapping clips, denoted as $\{v_k\}_{k=1}^K$. Considering the $i$-th and $j$-th video segments, where $j\geq i$, we can select any frame within the $i$-th segment as a potential starting time point $t_i$. Similarly, we can choose any frame from the $j$-th segment as a potential ending time point $t_j$. Thus, a candidate time period $T_{ij} = [t_i, t_j]$ is established. During feature extraction, each video segment ${v_k} (k = 1, ..., K)$ is individually processed through a C3D network\cite{tran2015learning} to extract per-frame features. Within each segment, average pooling is then applied over all the frame-level features to derive $\bm{f}_K\in {\mathbb{R}^{K\times {d_s}}}$ for that segment. For a given candidate moment $T_{ij} = [t_i, t_j]$, we perform max-pooling  operations over the sequence of segment features $\bm{f}_i, \bm{f}_{i+1}, ..., \bm{f}_j$ to aggregate them into a single feature vector $\bm{f}_{ij}^{c}$. This aggregated feature $\bm{f}_{ij}$ serves as the video features for the candidate moment $T_{ij}$. Finally, we assemble all the aggregated features $\bm{f}_{ij} (i, j = 1, ...,K; j \geq i)$ into a temporal feature map $\bm F\in {\mathbb{R}^{K\times K\times {d_v}}}$. This feature map encapsulates the temporal evolution of the video content and can be leveraged for downstream tasks such as moment retrieval.

\begin{table}[t] 
	\centering    
	\caption{lookup table for main notations in this paper}
	\label{tab1}
	\setlength{\tabcolsep}{1.5pt} 
	\begin{tabular}{ll}
		\hline
		\hline
		\textbf{Notations} & \textbf{Description} \\
		\hline
		$\bm{V}$ & A video corpus \\
		$\bm{S}$ & A text paragraph query \\
		$V_n$ & The $n$-th video in the video corpus \\
		$V^*$ & The video corresponding to the paragraph query \\
		$M$ & The number of sentences in a paragraph \\
		$Q$ & The number of candidate moments. \\
		$s_m$ & The $m$-th sentence \\
		$\bm{h}_m$ & The feature representation of the $m$-th sentence \\
		$t_s$ & The start times of the event \\
		$t_e$ & The end times of the event \\
		$\bm{f}_{cls,v}$ & The visual class token features \\
		$\bm{f}_{cls,t}$ & The textual class token features \\
		$B$ & The batch size \\
		$\bm{P}_m$ & The 2D score map of the $m$-th sentence. \\
		$\bm{F}$ & The temporal feature map \\
		$\left\{w_i\right\}_{i=1}^Q$ & The $Q$ balancing factors \\
		$\{{\bm{P}}_m^f\} _{m = 1}^M$ & The 2D score map of the forward branch \\
		$\{{\bm{P}}_m^r\} _{m = 1}^M$ & The 2D score map of the reverse branch \\
		${gt}_{m,ij}^f$ & The soft label of the forward branch \\
		${gt}_{m,ij}^r$ & The soft label of the reverse branch \\
		\hline
		\hline
	\end{tabular}
\end{table}

\subsection{Video-Paragraph Cross Modal Retrieval}

In the retrieval branch, our goal is to achieve precise cross-modal retrieval between videos and paragraphs and construct a coarse-grained feature space. Specifically, we introduce two learnable class tokens: ${\bm{f}}_{cls,v}$ to aggregate the global features of the video, and ${\bm{f}}_{cls,t}$ is used to integrate the global featuresn of the paragraph:

\begin{equation}
\begin{aligned}
	{\tilde{\bm{f}}}_{cls,v}= {\bm{E}_1}[\bm{f}_{cls, v},\bm{f}_1,\bm{f}_2,...\bm{f}_K],\\	
	{\tilde{\bm{f}}}_{cls,t}= {\bm{E}_2}[\bm{f}_{cls, t},\bm{h}_1,\bm{h}_2,...,\bm{h}_M],
\end{aligned}
\end{equation}
where `[]' denotes the concatenation operation, $\bm{E}_1$ and $\bm{E}_1$ are two Transformer\cite{vaswani2017attention} encoders that do not share parameters. ${\tilde{\bm{f}}}_{cls,v}$ and ${\tilde{\bm{f}}}_{cls,t}$ represent the aggregated global visual and textual class token features, respectively. Then, we use the InfoNCE loss\cite{he2020momentum} and  the triplet ranking loss\cite{schroff2015facenet}  as training objective for video-paragraph cross modal retrieval. Given a positive paragraph-video pair, the InfoNCE loss over a batch $B$ can be defined as:

\begin{equation}
\mathcal{L}_{{nce}}^{r} =  \mathcal{L}_{{nce}}^{t2v} + \mathcal{L}_{{nce}}^{v2t},
\end{equation}
\begin{equation}
\mathcal{L}_{{nce}}^{t2v} = - \frac{1}{B} \sum_{b = 1}^B \left[ \log \frac{e^{S^r(\bm{\tilde{f}}_{{cls},t}^b, \bm{\tilde{f}}_{{cls},v}^b) \cdot \lambda}}{\sum_{z = 1}^B e^{S^r(\bm{\tilde{f}}_{{cls},t}^b, \bm{\tilde{f}}_{{cls},v}^z) \cdot \lambda}} \right],
\end{equation}
\begin{equation}
\mathcal{L}_{{nce}}^{v2t} = - \frac{1}{B} \sum_{b = 1}^B \left[ \log \frac{e^{S^r(\bm{\tilde{f}}_{{cls},v}^b, \bm{\tilde{f}}_{{cls},t}^b) \cdot \lambda}}{\sum_{z = 1}^B e^{S^r(\bm{\tilde{f}}_{{cls},v}^b, \bm{\tilde{f}}_{{cls},t}^z) \cdot \lambda}} \right],
\end{equation}
where $\lambda$ is a learnable scaling parameter.  $\bm{\tilde{f}}_{{cls},t}^b$ and $\bm{\tilde{f}}_{{cls},v}^b$ denote the class token features  for the b-th pair corresponding of the paragraph and video features within a batch $B$, respectively. $S^r(\bm{\tilde{f}}_{{cls},t}^b, \bm{\tilde{f}}_{{cls},v}^b)$ is the cosine similarity between the class token features $\bm{\tilde{f}}_{{cls},t}^b$ and $\bm{\tilde{f}}_{{cls},v}^b$. The triplet ranking loss is defined as follows:
\begin{equation}
\mathcal{L}_{{trip}}^{r} =  \mathcal{L}_{{trip}}^{t2v} + \mathcal{L}_{{trip}}^{v2t},
\end{equation}
\begin{equation}
\mathcal{L}_{{trip}}^{t2v} = \frac{1}{B} \sum_{b=1}^B \left[ \max \left(0, \Delta + S^r(\bm{\tilde{f}}_{cls,t}^b, \bm{\tilde{f}}_{cls,v}^{b\,-}) - S^r\left( \bm{\tilde{f}}_{cls,t}^b, \bm{\tilde{f}}_{cls,v}^b \right)\right) \right],
\end{equation}
\begin{equation}
\mathcal{L}_{{trip}}^{v2t} = \frac{1}{B} \sum_{b=1}^B \left[ \max \left(0, \Delta + S^r(\bm{\tilde{f}}_{cls,t}^{b\,-}, \bm{\tilde{f}}_{cls,v}^b) - S^r\left( \bm{\tilde{f}}_{cls,t}^b, \bm{\tilde{f}}_{cls,v}^b \right)\right) \right],
\end{equation}
where $\Delta$ is the margin. Similarly, $\bm{\tilde{f}}_{{cls},t}^{b\,-}$ and $\bm{\tilde{f}}_{{cls},v}^{b\,-}$ represent the negative class token features of the paragraph and video, respectively. These negative class token features are sampled from other samples within the same batch corresponding to the given $\bm{\tilde{f}}_{{cls},t}^b$ and $\bm{\tilde{f}}_{{cls},v}^b$. The total cross-modal retrieval loss is expressed as $\mathcal{L}_{{cmr}} = \mathcal{L}_{{trip}}^{r} + \beta_1 \mathcal{L}_{{nce}}^{r} $, with $\beta_1$ serving as a balancing factor.

\subsection{Visual-Textual Consistency on LD}

In the grounding branch, our goal is to achieve precise cross-modal grounding between video segments and textual paragraphs while constructing a fine-grained feature space. To this end, we propose a new method that learns cross-modal semantic consistency across local, global, and temporal dimensions, thereby enhancing the model's grounding capabilities. In the local dimension (LD), we develop a cross-modal feature consistency learning network based on 2D-TAN\cite{zhangSongyang2020}. We denote the feature representation of the $m$-th sentence within a paragraph as $\bm{h}_m$ and the corresponding temporal feature map associated with that paragraph as $\bm{F}_m$. To construct our model, we fuse these features through Eq. (8), obtaining the fused temporal feature map:

\begin{equation}
{\tilde{\bm{F}}}_m(i,j,:) = {\bm W_s}{\bm{h}_{m}^T} \odot {\bm W_v}{\left( {{\bm{F}_m}(i,j,:)} \right)^T},
\end{equation}
where $\bm W_s$ and  $\bm W_v$ are two learnable parameters, $T$ denotes the transpose operation, and $\odot$ represents the Hadamard product. 

Since Eq. (8) operates between sentence features and individual candidate time features, it ignores the contextual relationship among adjacent candidate time features. To overcome this limitation, we introduce the Temporal Adjacent Network (TAN) and feed ${\tilde{\bm{F}}}_m$ into it. This network, consisting of four convolutional layers, captures the associative relationships across different candidate moments. To calculate the response scores for the $m$-th sentence, we pass the temporal feature map ${\bar{\bm{F}}}_m$, which is the output of the Temporal Attention Network (TAN), through a prediction layer (PL). This layer consists of a fully connected layer followed by a Sigmoid activation function. This process generates a 2D score map ${\bm{P}}_m = {(p_{m,ij})_{K \times K}}$. Here, $p_{m,ij}$ represents the probability score indicating the correlation between the candidate moment that starts at the $i$-th video segment and ends at the $j$-th video segment, and the $m$-th sentence. In a weakly supervised setting, inspired by SCN\cite{lin2020weakly}, we propose a method for reconstructing candidate moment features.

Specifically, given the $m$-th sentence in a paragraph, we first select the top $Q=3$ candidate moments with the highest probability scores based on the predicted ${\bm{P}}_m$. The corresponding features for these candidate moments are denoted as $\{ {\bm{F}}_m^q\} _{q = 1}^Q$, derived from the frame-level features $\bm{f}_K\in {\mathbb{R}^{K\times {d_s}}}$. We then sequentially feed the features $\{ {\bm{F}}_m^q\} _{q = 1}^Q$ of the $Q$ candidate moments and the masked sentence features $\bm{\hat{h}}_m$ into a language reconstructor to predict the masked word features. The language reconstructor employed in our method comprises a Transformer's encoder and decoder, alongside a single fully connected layer and a softmax function. This workflow can be formulated as follows:
\begin{equation}
{\bm{f}}_m^q = {\bm{D}}({{{\hat{\bm{ h}}}_m},{\bm{E}}({{\bm{F}}_m^q})}), q = 0,1,...,Q; m = 1,2,...,M,
\end{equation}
\begin{equation}
\left\{ {\bm{e}}_{m,j}^q\right\} _{j = 1}^J = \mathrm{Softmax}\left(\mathrm{FC}\left({\bm{f}}_m^q\right)\right),
\end{equation}
where $\bm E$ and $\bm D$ represent the Transformer encoder and decoder respectively, and ${\bm{e}}_{m,j}^q$ denotes the probability of a word from the vocabulary, predicted using the visual features of the $q$-th candidate moment corresponding to the $m$-th sentence, appearing at the $j$-th masked word position. To ensure the accuracy of the prediction, we adopt reconstruction loss and rank loss to update the network parameters. 

In our method, the reconstruction loss can be defined as:
\begin{equation}
\mathcal{L}_{rec} = - \frac{1}{M} \mathop \sum \limits_{m = 1}^M \left( {\frac{1}{Q}\mathop \sum \limits_{q = 1}^Q \left( {\mathop \sum \limits_{j = 1}^J \log p\left( {{\bm{q}}_{m,j}\mid {\bm{e}}_{m,j}^q} \right)} \right)} \right),
\end{equation}	
where ${\bm{q}}_{m,j}^{} \in {\mathbb{R}^{{n_v} \times 1}}$ is the label for the $j$-th word appearing in the $m$-th sentence. The rank loss used in our method is defined as:
\begin{equation}
\mathcal{L}_{rank} = -\frac{1}{M} \mathop \sum \limits_{m = 1}^M \left( {\frac{1}{Q}\mathop \sum \limits_{q = 1}^Q \left( {{R^q}\log \left( {\frac{{\exp \left( {p_m^q} \right)}}{{\mathop \sum \limits_{q = 1}^Q \exp \left( {p_m^q} \right)}}} \right)} \right)} \right),
\end{equation}
where ${p_m^q}$ represents the probability scores corresponding to the top $Q$ selected candidate moments. $R^q$ is the reward score set based on the number of candidate moments $Q$, where higher-ranked moments among the $Q$ candidates are assigned greater reward scores. Based on prior experience in \cite{lin2020weakly}, this paper sets the reward score for the top-ranked candidate moment as 1, and the reward scores for subsequent candidate moments decrease progressively with a step size of $1/Q-1$. Therefore, the total loss of visual-text feature consistency in the local dimension can be expressed as:
\begin{equation}
{\mathcal{L}_{local}} = \mathcal{L}_{rec}+ \mathcal{L}_{rank}.
\end{equation}

\begin{figure}[t!]
\centering
\includegraphics[width=0.90\linewidth]{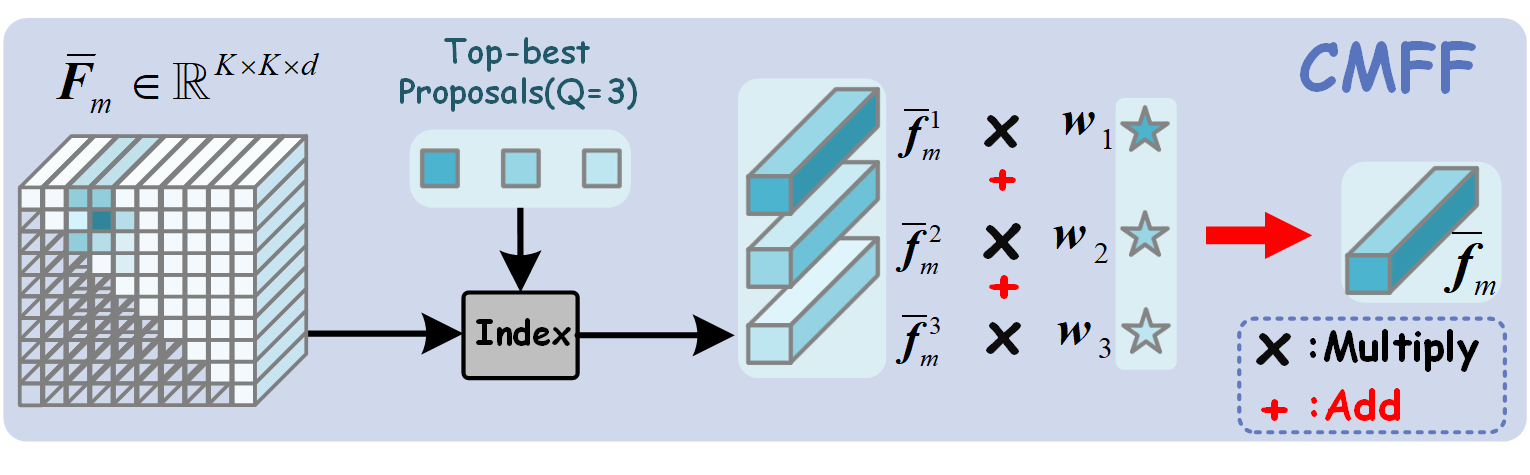}
\caption{Details of the candidate moments feature fusion(CMFF). Index refers to indexing the corresponding visual features in the fused temporal feature map ${\bar{\bm{F}}}_m$ based on the top Q candidate moments.}
\vspace{-14pt} 
\label{Fig04}
\end{figure}

\subsection{Visual-Textual Feature Alignment on GD}

On the global dimension (GD), we achieve cross-modal feature alignment by performing contrastive learning between videos and paragraphs. Specifically, for the $m$-th sentence, we select the top $Q=3$ candidate moments with the highest probability score according to $\bm{P}_m$ and input them along with the fused temporal feature map ${\bar{\bm{F}}}_m$ into the CMFF (as shown in Fig. \ref{Fig04}). Through the CMFF, we obtain the visual representation $\bar{\bm{f}}_m$ corresponding to the $m$-th sentence, which can be represented as follows:
\begin{equation}
\bar{\bm{f}}_m = \bar{\bm{f}}_m^1\ \times w_1 + \bar{\bm{f}}_m^2\ \times w_2 + \bar{\bm{f}}_m^3\ \times w_3,
\end{equation}
where $\left\{\bar{\bm{f}}_m^q\right\}_{q=1}^Q$ denotes the features corresponding to the candidate moments from the temporal feature map ${\bar{\bm{F}}}_m$. $\left\{w_i\right\}_{i=1}^Q$ are balancing factors used to weight features from different candidate moments. Compared to using only the features of the highest probability candidate moment as the visual representation of a sentence, the visual representations obtained through CMFF exhibit greater robustness and accuracy.

To align the global feature representations between paragraphs and their corresponding videos, we introduce a learnable class token $\bm{f}_{cls, v}^{\prime}$ to aggregate the visual features $\{\bar{\bm{f}_m}\}_{m=1}^M$ corresponding to each sentence in the paragraph query. The process is as follows:
\begin{equation}
\begin{aligned}
	\tilde{\boldsymbol{\bm{f}}}_{c l s, v}^{\prime} = {\bm{E}_3}[\bm{f}_{cls, v}^{\prime},\bar{\bm{f}}_1,\bar{\bm{f}}_2,...,\bar{\bm{f}}_M],
\end{aligned}
\end{equation}
where `[]' denotes the concatenation operation. $\bm{E}_3$ is a Transformer encoders that share parameters with $\bm{E}_1$. $\tilde{\boldsymbol{\bm{f}}}_{c l s, v}^{\prime} $ denotes the aggregated global visual class token features. Then we use the visual class token ${\tilde{\bm{f}}}_{cls,v}$ obtained from VPCMR as supplementary visual information. This token is fused with $\tilde{\boldsymbol{\bm{f}}}_{c l s, v}^{\prime} $. The operation is as follows:
\begin{equation}
\tilde{\bm{f}}_{cls,f} = \text{Conv}_{1 \times 1}([\tilde{\bm{f}}_{cls,v}, \tilde{\bm{f}}_{cls,v}^\prime]),
\end{equation}
where `[]' and $\text{Conv}_{1 \times 1}$ represent the concatenation and the ${1 \times 1}$  convolution operation, respectively. $\tilde{\bm{f}}_{cls,f}$ is the newly fused visual class token. It not only inherits the original global visual information but also integrates the visual information corresponding to each sentence in the paragraph query, thereby possessing stronger discriminative capabilities. To bring closer the distance between the features of the paragraph query and its corresponding video, and to push apart the features of unmatched videos. We follow the process in VPCMR to  jointly employ the InfoNCE loss $\mathcal{L}_{{nce}}^{g}$  and the triplet ranking loss $\mathcal{L}_{{trip}}^{g}$. These losses can be obtained by replacing $S^r(\bm{\tilde{f}}_{{cls},t}, \bm{\tilde{f}}_{{cls},v})$ with $S^g(\bm{\tilde{f}}_{{cls},t}, \bm{\tilde{f}}_{{cls},f})$ in Eq. (2) and (5), respectively. We denote the total loss for aligning text and visual features on the global dimension as $\mathcal{L}_{{global}} = \mathcal{L}_{{trip}}^{g} + \beta_2 \mathcal{L}_{{nce}}^{g} $, where $\beta_2$ is the weight of the InfoNCE loss.

To reinforce the retrieval branch through the grounding branch, we design the GRRM.  Specifically, we use the scores $S^g(\cdot)$ calculated through the grounding branch as pseudo-labels to supervise and constrain the scores $S^r(\cdot)$ of the retrieval branch. The loss is calculated as follows:
\begin{equation}
\mathcal{L}_{mse} = \frac{1}{|B|} \sum_{(i,j) \in B} \left(S^r(\bm{\tilde{f}}_{cls,t}^i, \bm{\tilde{f}}_{cls,v}^j) - S^g(\bm{\tilde{f}}_{cls,t}^i, \bm{\tilde{f}}_{cls,f}^j)\right)^2,
\end{equation}
where $B$ is the batch size. The scores ranges for $S^g(\cdot)$ and $S^r(\cdot)$ are between $-1$ and $1$. Through GRRM, we bring the coarse-grained feature space constructed by the retrieval branch closer to the fine-grained feature space obtained by the grounding branch. 
This facilitates  the global visual class token features ${\tilde{\bm{f}}}_{cls,v}$ and textual  class token features${\tilde{\bm{f}}}_{cls,t}$, used in the retrieval branch to develop strong discriminative capabilities, thereby achieving the goal of enhancing the retrieval branch through the grounding branch.

\begin{figure}[t]
	\centering
	\includegraphics[width=0.90\linewidth]{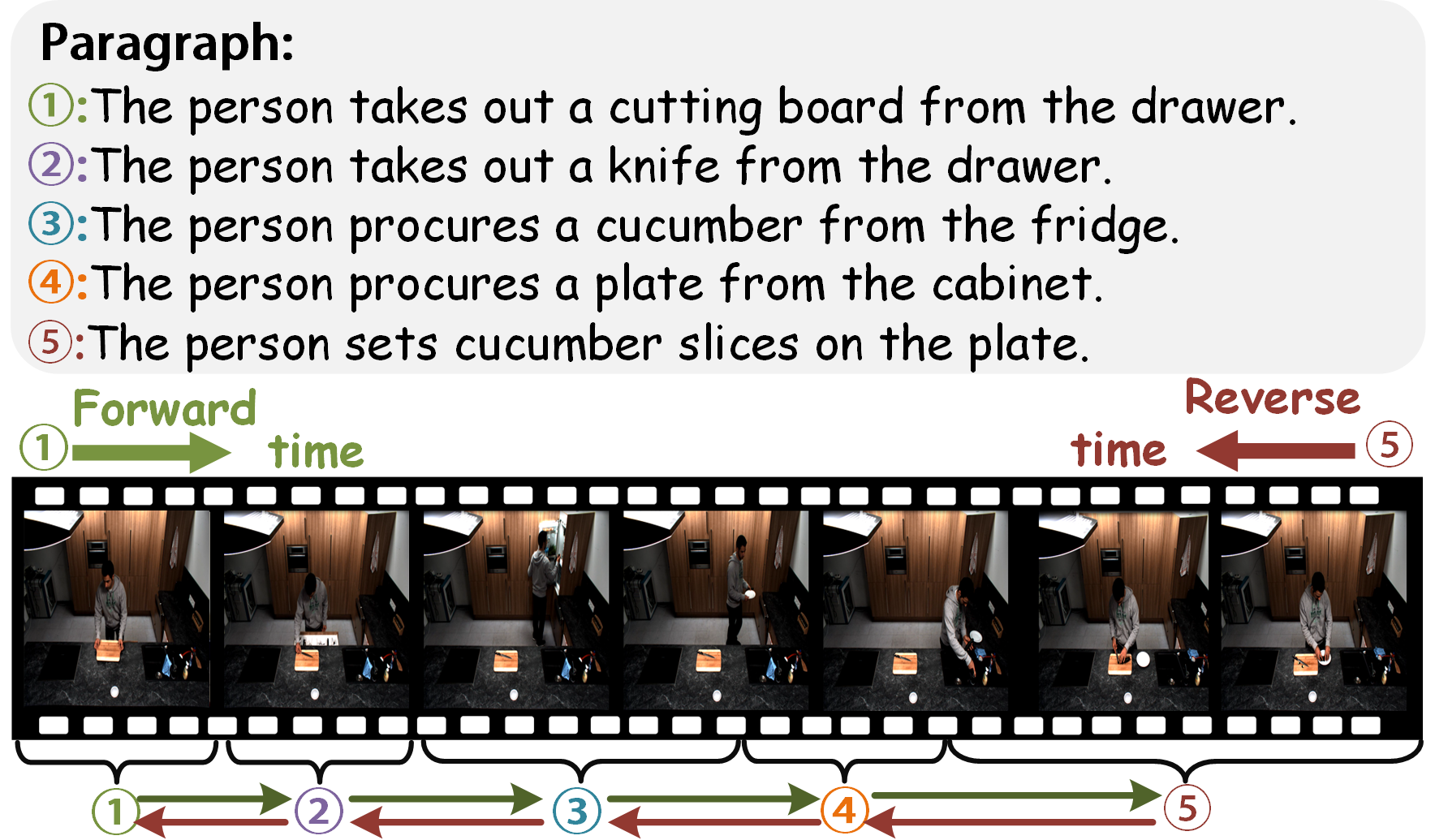}
	\caption{Diagram of the chronological order of events in a video.}
	\vspace{-13pt} 
	\label{Fig05}
\end{figure}

\subsection{Bidirectional Temporal Synchronization of Events on TD}

Events in videos typically occur in a specific chronological order, and similarly, sentences describing these events in text paragraphs follow a certain logical sequence (as shown in Fig. \ref{Fig05}). To achieve precise synchronization between video segments and their corresponding event descriptions in text, and thereby enhance the discriminative power of cross-modal features, we develop an effective model for visual-textual temporal synchronization. This model employs a bidirectional event-sentence alignment strategy to ensure tight synchronization and precise alignment between textual content and video segments. It takes into account contextual differences within both forward and reverse information flows, facilitating accurate matching of text to corresponding video segments.

\begin{figure}[t!]
\centering
\includegraphics[width=0.90\linewidth]{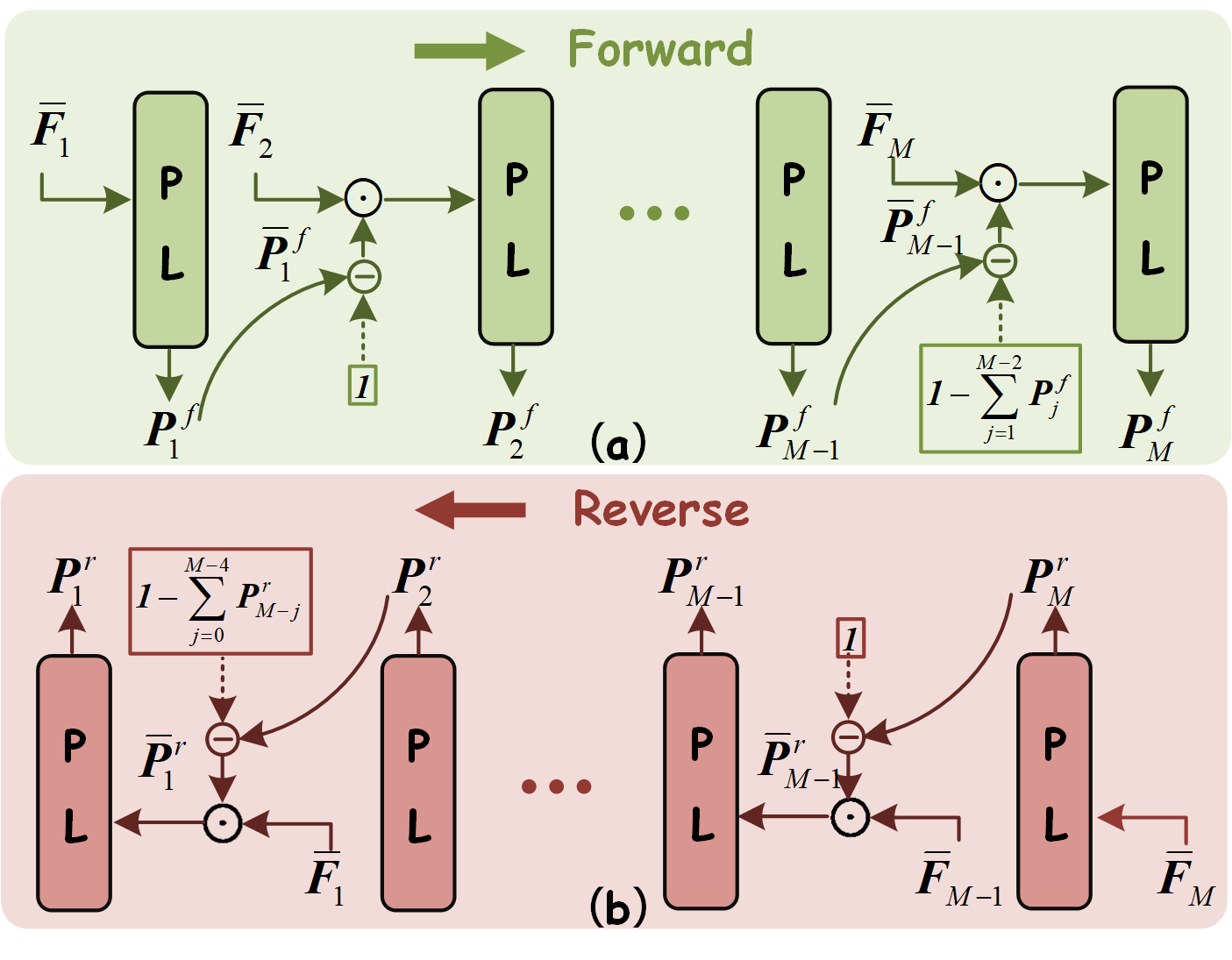}
\caption{Details of the bidirectional temporal synchronization of Events on TD (BTSE-TD).}

\label{Fig06}
\end{figure}

Our visual and text temporal synchronization model comprises two distinct branches: a forward branch and a reverse branch. Each branch is equipped with an independent prediction layer, consisting of a fully connected layer and a subsequent Sigmoid activation function. Notably, the parameters across these branches are not shared, ensuring their independence. The input to this model is a set of temporal feature maps $\{\bm{\bar {F}}_m\} _{m = 1}^M$, each generated by the TAN  network and corresponding to a sentence within the paragraph. In the forward synchronization process, we initiate by selecting the temporal feature map $\bm{\bar {F}}_1$ corresponding to the first sentence of the paragraph and feeding it into the prediction layer. This step yields the predicted candidate moment score map $\bm{P}_1^f \in {\mathbb{R}^{K \times K}}$ for that particular sentence. Subsequently, we perform an element-wise negation operation on $\bm{P}_1^f$ by calculating $\bm{1} - \bm{P}_1^f$, resulting in a new map $\bm{\bar{P}}_1^f$. We then proceed with an element-wise multiplication of $\bm{\bar {P}}_1^f$ and $\bm{\bar {F}}_2$. The outcome of this multiplication is once again fed into the prediction layer to derive the candidate moment score map for the second sentence. This iterative process continues in a similar manner until the $M$-th sentence of the paragraph is processed. The entire process can be visually represented in Fig. \ref{Fig06}.

It is worth noting that the candidate moment score map for the $m$-th sentence is iteratively computed, leveraging the results from the processing of the preceding sentence. This process can be formulated as:
\begin{equation}
{\bm{P}}_{m}^f = {\bm{P}}{{\bm{L}}^f}\left( {{\bm{\bar {F}}}_m \odot {\rm{expand}}\left( {{\bm{1}} - \sum\limits_{j = 1}^{m - 1} {{\bm{P}}_j^f} } \right)} \right), {\rm{1}} < m \le M,
\end{equation}
where ``expand'' refers to a channel expansion operation that increases the dimensionality of the score map from $K\times K$ to $K\times K\times d$, $\odot $ denotes the Hadamard product, and ${\bm{P}}{{\bm{L}}^f}$ represents the prediction layer. This approach effectively highlights the predicted results specific to the current sentence, mitigating the influence of preceding sentences on the current one. As a result, it achieves effective alignment between the current sentence and its corresponding video segment, facilitating forward temporal synchronization between visual and corresponding textual features on the temporal dimension.

In the reverse synchronization branch, we begin with the final sentence (specifically, the $M$-th sentence) in the paragraph. The temporal  feature map $\bm{\bar {F}}_M$ associated with this sentence is inputted into the prediction layer, generating a candidate moment score map ${\bm{P}}_M^r$ solely based on its content. Then, this prediction result is used to participate in the calculation of the candidate moment score map ${\bm{P}}_{M-1}^r$ corresponding to the $M-1$-th sentence. Iteratively, we derive a prediction score map for each sentence within the paragraph. This sequential approach enables fine-grained predictions, progressing backward to forward through time. The entirety of this process can be formalized as follows:
\begin{equation}
\bm{P}_{m}^{r} = {\bm{P}}{{\bm{L}}^r}\left( {{\bm{\bar {F}}}_m\odot {\rm{expand}}\left( {{\bm{1}} - \sum\limits_{j = 0}^{M - m - 1} {{\bm{P}}_{M - j}^r} } \right)} \right), {\rm{1}} \le m < M,
\end{equation}
where ${\bm{P}}{{\bm{L}}^r}$ denotes the prediction layer within the reverse synchronization branch. Notably, the prediction layers in the forward and reverse branches are parameterized independently, without shared parameters. The alignment of video and text, both in forward and reverse directions, effectively attains temporal synchronization between visual and textual features. This synchronization plays a crucial role in enhancing the performance of grounding tasks, ensuring accurate alignment and improved comprehension on the temporal dimension.

Additionally, compared to the score maps $\{ {{\bm{P}}_m}\} _{m = 1}^M$ of candidate moments generated without the aid of this module, the score maps $\{{\bm{P}}_m^f\} _{m = 1}^M$ and $\{{\bm{P}}_m^r\} _{m = 1}^M$
predicted by this module exhibit more effective temporal synchronization between video segments and text, facilitating a finer-grained correspondence between visual and textual elements. Taking this into account, we select the candidate moments with the highest predicted probability scores from the score map $\{{\bm{P}}_m^f\} _{m = 1}^M$ to form one set of pseudo-labels. Likewise, from the score map $\{{\bm{P}}_m^r\} _{m = 1}^M$, we choose the candidate moments with the highest probability scores to constitute another set of pseudo-labels. These two sets of pseudo-labels, which represent the most optimal candidate moments predicted in both directions, collaboratively contribute to the optimization of the overall score map $\{ {{\bm{P}}_m}\} _{m = 1}^M$.

Specifically, for the forward synchronization branch, we select the candidate moment with the highest probability score, denoted as $p_{m,ab}^f$, from the score map ${\bm{P}}_{m}^f=({p_{m,ij}^{f}})_{K \times K}$ of the $m$-th sentence. Subsequently, we designate the candidate moment corresponding to $p_{m,ab}^f$ as the pseudo label for the $m$-th sentence. In this study, we adopt a methodology similar to 2D-TAN to generate soft label ${gt}_{m,ij}^f$ and ${gt}_{m,ij}^r$ that constrain the score map $\{\bm{P}_m\}_{m = 1}^M$ of candidate moments. For the forward synchronization branch, the following the binary cross loss function is employed to optimize the alignment between the predicted candidate moments and these soft labels:

\begin{align}
	{\mathcal{L}^f} =  - \sum_{m=1}^M \sum_{(i,j)\in {\bm{C}}} \Big( & {gt}_{m,ij}^f \log(p_{m,ij}) \nonumber \\
	& + (1 - {gt}_{m,ij}^f)\log(1 - p_{m,ij}) \Big),
	\end{align}
where $\bm{C}=\{(i,j)|1\leq i \leq j \leq K\}$. Similarly, for the reverse synchronization branch, the binary cross loss function used in this paper can be expressed as

\begin{align}
	{\mathcal{L}^r} = - \sum_{m=1}^M \sum_{(i,j) \in \bm{C}} & \left( {gt}_{m,ij}^r \log(p_{m,ij}) \right. \nonumber \\
	& \left. + (1 - {gt}_{m,ij}^r) \log(1 - p_{m,ij}) \right).
	\end{align}
	The total loss of bidirectional temporal synchronization of events on temporal can be expressed as:
	\begin{equation}
{\mathcal{L}_{time}} = {\mathcal{L}^f} + {\mathcal{L}^r}.
\end{equation}

\subsection{Training and Inference}
\subsubsection{Training}
The entire network is trained end-to-end, with model parameters optimized through the following total loss:

\begin{equation}
{\mathcal{L}_{total}} = \mathcal{L}_{cmr}+ \mathcal{L}_{local}+ \mathcal{L}_{global}+ \mathcal{L}_{time}+ \mathcal{L}_{mse}.
\end{equation}
\subsubsection{Inference}
During the inference process, we first use VPCMR to calculate the similarity scores between all videos in the video corpus and the paragraph query, selecting the video with the highest similarity as the retrieval result. Then, using the VTC-LD module, we predict the start and end times of the events corresponding to each sentence in the paragraph within the retrieved video. It is important to note that TSVT-TD and VTFA-GD do not participate in retrieval and grounding during the testing phase.

\begin{table*}[h]
\centering
\caption{Comparison of the proposed method with state-of-the-art methods on the ActivityNet Captions and Charades-STA datasets. Evaluation results for R@10 IoU=0.3, R@100 IoU=0.3, R@10 IoU=0.5, R@100 IoU=0.5, R@10 IoU=0.7, and R@100 IoU=0.7 are listed. ``--'' indicates no reported data. Best and runner-up values are in bold and underlined, respectively.}
\label{Table1}
\small
\setlength{\tabcolsep}{4.5pt} 
\renewcommand{\arraystretch}{1.4}
\begin{tabular}{c|cc|cc|cc||cc|cc|cc}
	\hline
	\hline
	\multirow{3}{*}{Method} & \multicolumn{6}{c||}{ActivityNet Caption} & \multicolumn{6}{c}{Charades-STA} \\
	\cline{2-13}
	& \multicolumn{2}{c|}{IoU=0.3} & \multicolumn{2}{c|}{IoU=0.5} & \multicolumn{2}{c||}{IoU=0.7} & \multicolumn{2}{c|}{IoU=0.3} & \multicolumn{2}{c|}{IoU=0.5} & \multicolumn{2}{c}{IoU=0.7} \\
	& R@10 & R@100 & R@10 & R@100 & R@10 & R@100 & R@10 & R@100 & R@10 & R@100 & R@10 & R@100 \\
	\hline
	MCN\cite{anne2017localizing,chen2023joint} & -- & -- & 0.18 & 1.26 & 0.09 & 0.70 & -- & -- & 0.52 & 2.96 & 0.31 & 1.75 \\
	CAL\cite{victor2019temporal,chen2023joint} & -- & -- & 0.21 & 1.58 & 0.10 & 0.90 & -- & -- & 0.75 & 4.39 & 0.42 & 2.78 \\
	XML\cite{lei2020tvr,chen2023joint} & 3.21 & 12.48 & 1.69 & 7.58 & 0.10 & 0.90 & 0.70 & 2.47 & 0.32 & 1.42 & 0.16 & 0.78 \\
	HMAN\cite{paul2021text,chen2023joint} & -- & -- & 0.66 & 4.75 & 0.32 & 2.27 & -- & -- & 1.40 & 7.79 & 1.05 & 4.69 \\
	ReLoCLNet\cite{zhang2021video,chen2023joint} & 4.82 & 15.80 & 3.01 & 11.22 & 1.47 & 6.30 & 1.51 & 3.28 & 0.94 & 2.26 & 0.59 & 1.21 \\
	MS-SL\cite{dong2022partially,chen2023joint} & 10.80 & 28.31 & 5.85 & 15.65 & 2.46 & 6.60 & 4.46 & 17.61 & 2.55 & 10.05 & 0.91 & 3.76 \\
	JSG\cite{chen2023joint} & \underline{13.27} & \underline{40.61} & \underline{8.76} & \underline{29.98} & \underline{3.83} & \underline{15.78} & \underline{7.23} & \underline{28.71} & \underline{5.67} & \underline{22.50} & \underline{3.28} & \underline{12.34} \\
	\hline
	\textbf{DMR-JRG(Ours)} & \textbf{16.19} & \textbf{55.54} & \textbf{14.05} & \textbf{48.04} & \textbf{8.65} & \textbf{29.22} & \textbf{11.39} & \textbf{42.07} & \textbf{10.24} & \textbf{30.49} & \textbf{6.72} & \textbf{19.64} \\
	\hline
	\hline
\end{tabular}
\vspace{-10pt} 
\end{table*}

\begin{table}[t]
\centering
\caption{Compare the proposed method with state-of-the-art method on the TaCoS dataset. Evaluation results for R@10 IoU=0.3, R@100 IoU=0.3, R@10 IoU=0.5, R@100 IoU=0.5, R@10 IoU=0.7, and R@100 IoU=0.7 are listed. Best and runner-up values are in bold and underlined, respectively. “JSG*”shows our replicated results on TaCoS dataset.}
\label{Table2} 
\small 
\setlength{\tabcolsep}{2.2pt} 
\renewcommand{\arraystretch}{1.4} 
\begin{tabular}{c|cc|cc|cc}
	\hline
	\hline
	\multirow{2}{*}{Method} & \multicolumn{2}{c|}{IoU=0.3} & \multicolumn{2}{c|}{IoU=0.5} & \multicolumn{2}{c}{IoU=0.7} \\
	
	& R@10 & R@100 & R@10 & R@100 & R@10 & R@100 \\
	\hline
	JSG*\cite{chen2023joint} & \underline{7.23} & \underline{28.71} & \underline{5.67} & \underline{22.50} & \underline{3.28} & \underline{12.34} \\
	\hline 
	\textbf{DMR-JRG(Ours)} & \textbf{18.74} & \textbf{45.45} & \textbf{10.00} & \textbf{26.78} & \textbf{5.83} & \textbf{15.59} \\
	\hline
	\hline
\end{tabular}
\vspace{-8pt} 
\end{table}

\section{Experiments}
\subsection{Datasets and Evaluation Protocol}
\textbf{Datasets.} The proposed method is evaluated on video paragraph grounding datasets, including ActivityNet Captions\cite{caba2015activitynet}, Charades-STA\cite{rohrbach2012script} and TaCoS\cite{regneri2013grounding}. These datasets consists of videos and their corresponding paragraph queries.

\textbf{ActivityNet Captions.} This dataset contains $14,926$ untrimmed videos and successfully builds $19,811$ video-paragraph pairs based on these videos. It is noteworthy
that some of the videos are associated with multiple text paragraphs. The average duration of the videos is $117.60$ seconds,  while each text paragraph contains an average of $3.63$ sentences. In addition, the dataset has been divided into three subsets: training set, validation set$1$, and validation set$2$, containing $10,009$, $4,917$, and $4,885$  video-paragraph pairs respectively. In order to be consistent with previous research, we choose to treat validation set$2$ as the test set for subsequent evaluation.

\textbf{Charades-STA.} This dataset includes 6,672 videos of indoor activities. We follow the official data split, where the training set and the testing set contain 5,338 and 1,334 video-paragraph pairs, respectively. Additionally, the average length of the videos in the dataset is $29.8$ seconds, and text paragraphs average $2.41$ sentences.

\textbf{TaCoS.} This dataset combines the MPII corpus\cite{rohrbach2012script} with kitchen scene videos, specifically constructing $127$ videos focused on cooking activities. Each video corresponds to multiple text paragraphs, forming video-paragraph pairs. For the training, validation, and test sets, there are $1,107$, $418$, and $380$ such pairs, respectively. Additionally, the average length of the videos in the dataset is $4.79$ minutes, and text paragraphs average $8.75$ sentences.

\textbf{Evaluation Protocol.} In the VSRG task, the JSG\cite{chen2023joint} uses Recall@K IoU=m as an evaluation metric, abbreviated as R@K IoU=m. Specifically, R@K IoU=m refers to the percentage of test samples for which the Intersection over Union (IoU) between the predicted moments in the top K retrieved videos and the ground truth exceeds m, given a sentence query. For ease of comparison with previous VSRG methods, although our model uses paragraphs as queries during testing, the metric is calculated in the same way as JSG, based on individual sentences.

\subsection{Implementation Details}
Similar to the 2D-TAN\cite{zhangSongyang2020} method, this paper utilizes a pre-trained C3D model\cite{tran2015learning} on the UCF101 dataset\cite{karpathy2014large} for video feature extraction. Furthermore, it employs word2vec\cite{mikolov2013distributed} to obtain word embedding. The training process comprises 100 epochs, and throughout this process, the Adam optimizer is used with an initial learning rate of 0.0001 to update network parameters. Additionally, a learning rate decay mechanism is incorporated, where the rate is reduced by a factor of 0.8 every 20 epochs. During the training process, we use batch sizes of 16, 16, and 6 for the ActivityNet Captions, Charades-STA, and TaCoS datasets, respectively. Meanwhile, the margin $\Delta$ is set to 0.2, and the balancing factor $\beta_1$ and $\beta_2$ are set to 0.04 for all datasets. In addition, following the setup in 2D-TAN \cite{zhangSongyang2020}, we have configured the 2D score map size to $32 \times 32$ for the ActivityNet Caption dataset, and $16 \times 16$ for both the Charades-STA and TaCoS datasets.

\begin{table*}[t]
\centering
\caption{Ablation studies on ActivityNet Captions dataset. Evaluation results for R@10 IoU=0.3, R@100 IoU=0.3, R@10 IoU=0.5, R@100 IoU=0.5, R@10 IoU=0.7, and R@100 IoU=0.7 are listed. ``Base'' denotes the baseline method, ``Forw'' denotes the forward branches and ``Reve'' denotes the reverse branches. Best and runner-up values are in bold and underlined, respectively.}
\small
\setlength{\tabcolsep}{4.0pt} 
\renewcommand{\arraystretch}{1.4}
\label{Table3}
\begin{tabularx}{\textwidth}{c|c|c|c|c|c|c|c|cc|cc|cc}
	\hline
	\hline
	\multirow{2.3}{*}{Base} & \multirow{2.3}{*}{VPCMR} & \multirow{2.3}{*}{\makecell{VTC\\-LD}} & \multirow{2.3}{*}{\makecell{BTSE-TD\\ w/o Reve}} & \multirow{2.3}{*}{\makecell{BTSE-TD\\ w/o Forw}} & \multirow{2.3}{*}{\makecell{BTSE\\-TD}} &\multirow{2.3}{*}{\makecell{VTFA-GD\\ w/o GRRM}} & \multirow{2.3}{*}{\makecell{VTFA-GD\\ w/ GRRM}} & \multicolumn{2}{c|}{IoU=0.3} & \multicolumn{2}{c|}{IoU=0.5} & \multicolumn{2}{c}{IoU=0.7} \\

	& & & & & & & & R@10 & R@100 & R@10 & R@100 & R@10 & R@100 \\
	\hline
	$\checkmark$ & & & & & & & & 12.68 & 42.18 & 8.37 & 33.54 & 3.05 & 18.41 \\
	
	$\checkmark$ & $\checkmark$ & & & & & & & 13.16 & 43.27 & 8.96 & 35.87 & 3.72 & 19.83 \\
	
	$\checkmark$ & $\checkmark$ & $\checkmark$ & & & & & & 13.97 & 45.39 & 9.52 & 38.14 & 4.45 & 21.47 \\
	
	$\checkmark$ & $\checkmark$ & $\checkmark$ & $\checkmark$ & & & & & 14.69 & 48.47 & 12.38 & 42.29 & 5.27 & 20.73 \\
	
	$\checkmark$ & $\checkmark$ & $\checkmark$ &  & $\checkmark$ & & & & 14.35 & 49.14 & 11.82 & 42.56 & 5.96 & 22.51 \\
	
	$\checkmark$ & $\checkmark$ & $\checkmark$ & & & $\checkmark$ & & & 15.22 & 51.07 & 13.02 & 43.51 & 6.79 & 25.82 \\
	
	$\checkmark$ & $\checkmark$ & $\checkmark$ &  & & $\checkmark$ & $\checkmark$ & & \underline{15.76} & \underline{52.81} & \underline{13.34} & \underline{45.23} & \underline{7.21} & \underline{27.54} \\
	
	$\checkmark$ & $\checkmark$ & $\checkmark$ &  & & $\checkmark$ & & $\checkmark$ & \textbf{16.19} & \textbf{55.54} & \textbf{14.05} & \textbf{48.04} & \textbf{8.65} & \textbf{29.22} \\
	\hline
	\hline
\end{tabularx}
\end{table*}

\begin{table}[h]
\centering
\vspace{-14pt} 
\caption{Evaluate the impact of different configurations  of $\left\{w_i\right\}_{i=1}^Q$ on model performance on the ActivityNet Captions dataset. “Learnable” indicates that $\left\{w_i\right\}_{i=1}^Q$ are set as learnable parameters. Best and runner-up values are in bold and underlined, respectively.}
\label{Table4} 
\small 
\setlength{\tabcolsep}{2.6pt} 
\renewcommand{\arraystretch}{1.4} 
\begin{tabular}{c|cc|cc|cc}
	\hline
	\hline
	\multirow{2}{*}{Method} & \multicolumn{2}{c|}{IoU=0.3} & \multicolumn{2}{c|}{IoU=0.5} & \multicolumn{2}{c}{IoU=0.7} \\
	
	& R@10 & R@100 & R@10 & R@100 & R@10 & R@100 \\
	\hline
	$\{0.8,0.1,0.1\}$ & 15.56 & 54.32 & 13.45 & 46.89 & 7.50 & 27.23 \\
	
	$\{0.6,0.2,0.2\}$ & 15.84 & 54.87 & 13.77 & 47.32 &  8.19 & 27.91 \\
	
	$\{0.5,0.25,0.25\}$ & 15.97 & 55.02 & \underline{14.02} & 47.76 & 8.03 & 28.13 \\
	
	$\{0.4,0.3,0.3\}$  & \textbf{16.19} & \textbf{55.54} & \textbf{14.05} & \textbf{48.04} & \textbf{8.65} & \textbf{29.22}\\
	$\{0.33,0.33,0.33\}$  & \underline{16.03} & \underline{55.11} & 13.74 & 47.59 & 8.17 &\underline{29.08} \\
	\hline 
	Learnable & 15.94 & 54.23 & 13.29 & \underline{47.92} & \underline{8.28} & 28.51 \\
	\hline
	\hline
\end{tabular}
\vspace{-8pt} 
\end{table}

\subsection{ Comparison with State-of-the-Art Methods}
In this section, we evaluate our proposed method against state-of-the-art methods on three public datasets, as shown in Tables II-III. Additionally, due to the lack of existing research on VPRG task, to validate the effectiveness of our approach, we choose to compare it with existing VSRG methods. It is important to note that although VSRG methods use a single sentence as a query, our proposed DMR-JRG method utilizes paragraphs composed of multiple sentences for querying. Both approaches employ the same datasets, such as ActivityNet Captions, Charades-STA, and TaCoS, during the training and testing phase, ensuring relative fairness. Additionally, the results of methods MCN\cite{anne2017localizing}, CAL\cite{victor2019temporal}, XML\cite{lei2020tvr}, HMAN\cite{paul2021text}, ReLoCLNet\cite{zhang2021video}, and MS-SL\cite{dong2022partially} on  ActivityNet Captions and Charades-STA datasets, as presented in Table~\ref{Table1}, were replicated by JSG\cite{chen2023joint}.

\textbf{Results on ActivityNet Captions dataset:} Table~\ref{Table1} presents a performance comparison of our DMR-JRG method with current state-of-the-art VSRG methods on the ActivityNet Captions dataset. The results indicate that DMR-JRG outperforms existing methods across all evaluation metrics. Specifically, compared to the latest JSG\cite{chen2023joint} method, our approach shows significant improvements in the VSRG task, with increases of 2.92\% at R@10 IoU=0.3, 14.93\% at R@100 IoU=0.3, 5.29\% at R@10 IoU=0.5, 18.06\% at R@100 IoU=0.5, 4.82\% at R@10 IoU=0.7, and 13.44\% at R@100 IoU=0.7. These significant improvements not only demonstrate the effectiveness of our proposed method but also highlight the distinct advantages of paragraph-level over sentence-level retrieval and grounding. Moreover, the superior performance of DMR-JRG is largely attributed to the design of our dual-task reinforcing framework. By facilitating mutual promotion between the retrieval and grounding branches, we achieve more accurate cross-modal matching and grounding.

\begin{figure}[t]
\centering

\includegraphics[width=0.60\linewidth]{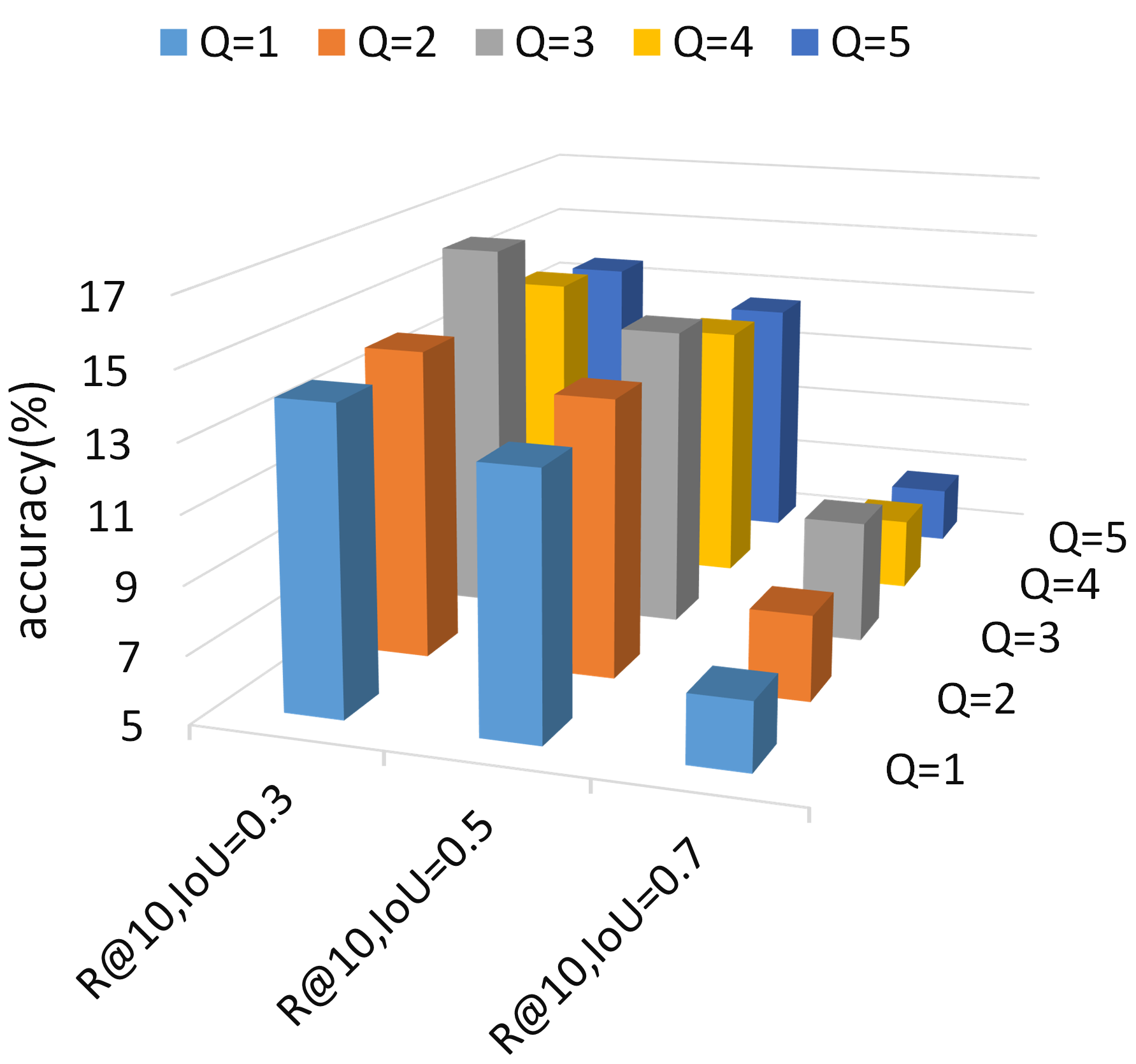}
\caption{Visualization of the impact of different Q values on model performance in the ActivityNet Captions dataset.}
\vspace{-14pt} 
\label{Fig07}
\end{figure}

\begin{figure*}[t]
\centering

\includegraphics[width=1\linewidth]{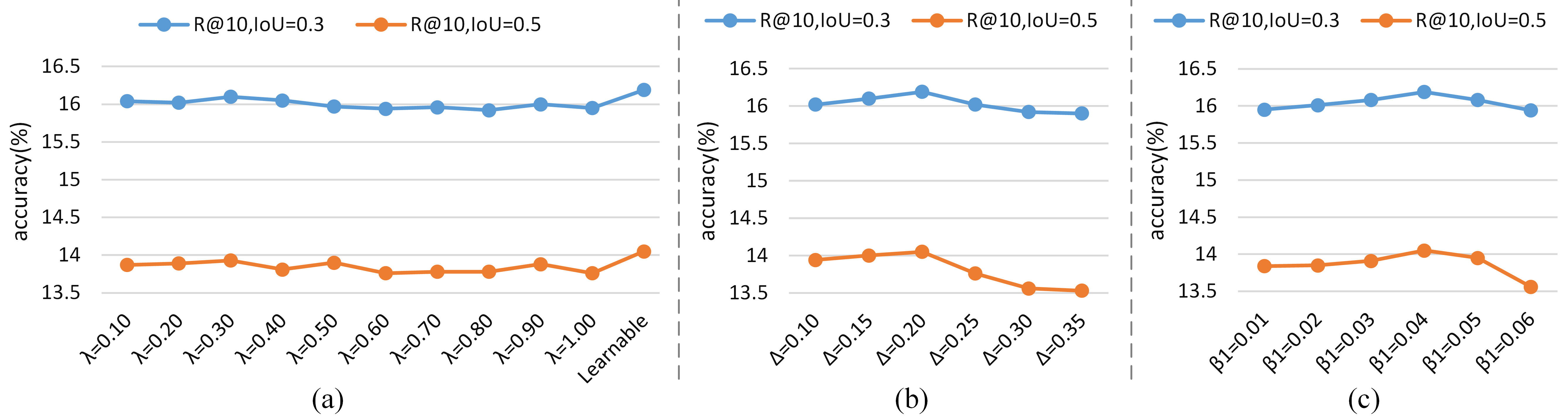}

\caption{Impact of different values of  $\lambda$, $\Delta$ and $\beta_1$ on model performance in the ActivityNet Captions dataset. “Learnable” indicates that $\lambda$ is treated as a learnable parameter.}

\label{Fig08}
\end{figure*}
\textbf{Results on Charades-STA dataset:} Table~\ref{Table1} presents a performance comparison of our DMR-JRG method with the current state-of-the-art VSRG methods on the Charades-STA dataset. Notably, the proposed method outperforms existing methods across all evaluation metrics. Specifically, the accuracies of our proposed method at R@10 IoU=0.3, R@100 IoU=0.3, R@10 IoU=0.5, R@100 IoU=0.5, R@10 IoU=0.7, and R@100 IoU=0.7 are 11.39\%, 42.07\%, 10.24\%, 30.49\%, 6.72\%, and 19.64\%, respectively. Compared to the second-best method JSG\cite{chen2023joint}, the proposed method shows improvements of 4.16\%, 13.36\%, 4.57\%, 7.99\%, 3.44\%, and 7.30\% in these metrics, respectively. These results amply demonstrate the versatility and effectiveness of the proposed method.

\textbf{Results on TaCoS dataset:} Table~\ref{Table2} shows a performance comparison between our method DMR-JRG and JSG\cite{chen2023joint} on the TaCoS dataset. Events in the TaCoS dataset all occur within the same kitchen scene, with minimal variations between them, presenting significant challenges for event grounding. Despite these challenges, our method significantly outperforms JSG. Specifically, our method achieves notable improvements across all evaluation metrics of 11.51\%, 16.74\%, 4.33\%, 4.28\%, 2.55\%, and 3.25\%, respectively. These results not only further confirm the effectiveness of the proposed method but also demonstrate that excavating the multi-dimensional consistency of local, global, and temporal dimensions between video segments and text paragraphs is crucial for achieving precise cross-modal matching and grounding.

\vspace{-5pt} 

\subsection{Ablation Studies}
Our proposed method DMR-JRG primarily consists of four modules: VPCMR, VTC-LD, BTSE-TD and VTFA-GD. To assess the individual contributions of these modules, we conduct ablation studies on the ActivityNet Captions dataset, as shown in Table~\ref{Table3}. In these experiments, we use the SCN \cite{lin2020weakly} from the VSG task as our baseline model, which is referred to as ``Base''. Moreover, we modify the input by replacing single sentences with paragraphs to capture richer contextual information. Finally, we train the model using cross-modal retrieval loss and visual-textual feature consistency loss to ensure fine-tuning and optimization of the baseline.

\textbf{Effectiveness of VPCMR.} To validate the effectiveness of the VPCMR module, we integrate this module into the baseline, forming the ``Base+VPCMR'' method. The results presented in Table~\ref{Table3} indicate that incorporating VPCMR significantly enhances model performance. Notably, on the ActivityNet Captions dataset, all evaluation metrics increase by 0.48\%, 1.09\%, 0.59\%, 2.33\%, 0.67\%, and 1.42\%, respectively. These findings clearly confirm the efficacy of the VPCMR module.

\textbf{Effectiveness of VTC-LD.} To validate the effectiveness of the VTC-LD module, we add it to the ``Base+VPCMR'' model. According to the experimental results in Table~\ref{Table3}, dataset, ``Base+VPCMR+VTC-LD'' shows improvements over ``Base+VPCMR'' by 0.81\%, 2.12\%, 0.56\%, 2.27\%, 0.73\%, and 1.64\% across all evaluation metrics. These results fully demonstrate the efficacy of the VTC-LD module.

\textbf{Effectiveness of BTSE-TD.} To validate the effectiveness of the BTSE-TD module, we add it to the ``Base+VPCMR+VTC-LD'' model, forming the ``Base+VPCMR+VTC-LD+BTSE-TD'' method. To further investigate the role of bidirectional temporal synchronization, we split this integrated method into two variants: ``Base+VPCMR+VTC-LD+(BTSE-TD w/o Reve)'', which only includes the forward branch, and ``Base+VPCMR+VTC-LD+(BTSE-TD w/o Forw)'', which only includes the reverse branch. The experimental results shown in Table~\ref{Table3} indicates that unidirectional synchronization shows certain effectiveness, while the simultaneous use of both forward and reverse branches significantly enhances performance.It is worth noting that on the ActivityNet Captions dataset, ``Base+VPCMR+VTC-LD+BTSE-TD'' improved by 1.25\%, 5.68\%, 3.50\%, 5.37\%, 2.34\%, and 4.35\% in all evaluation metrics respectively compared to ``Base+VPCMR+VTC-LD''. This highlights the importance of bidirectional temporal synchronization. Moreover, these results fully demonstrate that deeply exploring the temporal dimension consistency between video segments and text paragraphs within the grounding branches significantly boosts model performance.

\textbf{Effectiveness of VTFA-GD.} To validate the effectiveness of the VTFA-GD module, we incorporate it into the ``Base+VPCMR+VTC-LD+BTSE-TD'' model. According to the experimental results in Table~\ref{Table3}, ``Base+VPCMR+VTC-LD+BTSE-TD+(VTFA-GD w/ GRRM)'' improved upon ``Base+VPCMR+VTC-LD+BTSE-TD'' by 0.97\%, 4.47\%, 1.03\%, 4.53\%, 1.86\%, and 3.40\% across all evaluation metrics. Furthermore, to validate the effectiveness of the GRRM module, we remove GRRM from the VTFA-GD module, forming the ``Base+VPCMR+VTC-LD+BTSE-TD+(VTFA-GD w/o GRRM)''. Table~\ref{Table3} shows that the model with the GRRM module outperforms the one without it across all metrics, particularly with improvements of 2.73\% and 1.68\% at R@100 IoU=0.3 and R@100 IoU=0.7, respectively. This demonstrates the reinforcing effect of the grounding branch for the retrieval branch.

\begin{figure*}[h]
	\centering
	\includegraphics[width=0.85\textwidth]{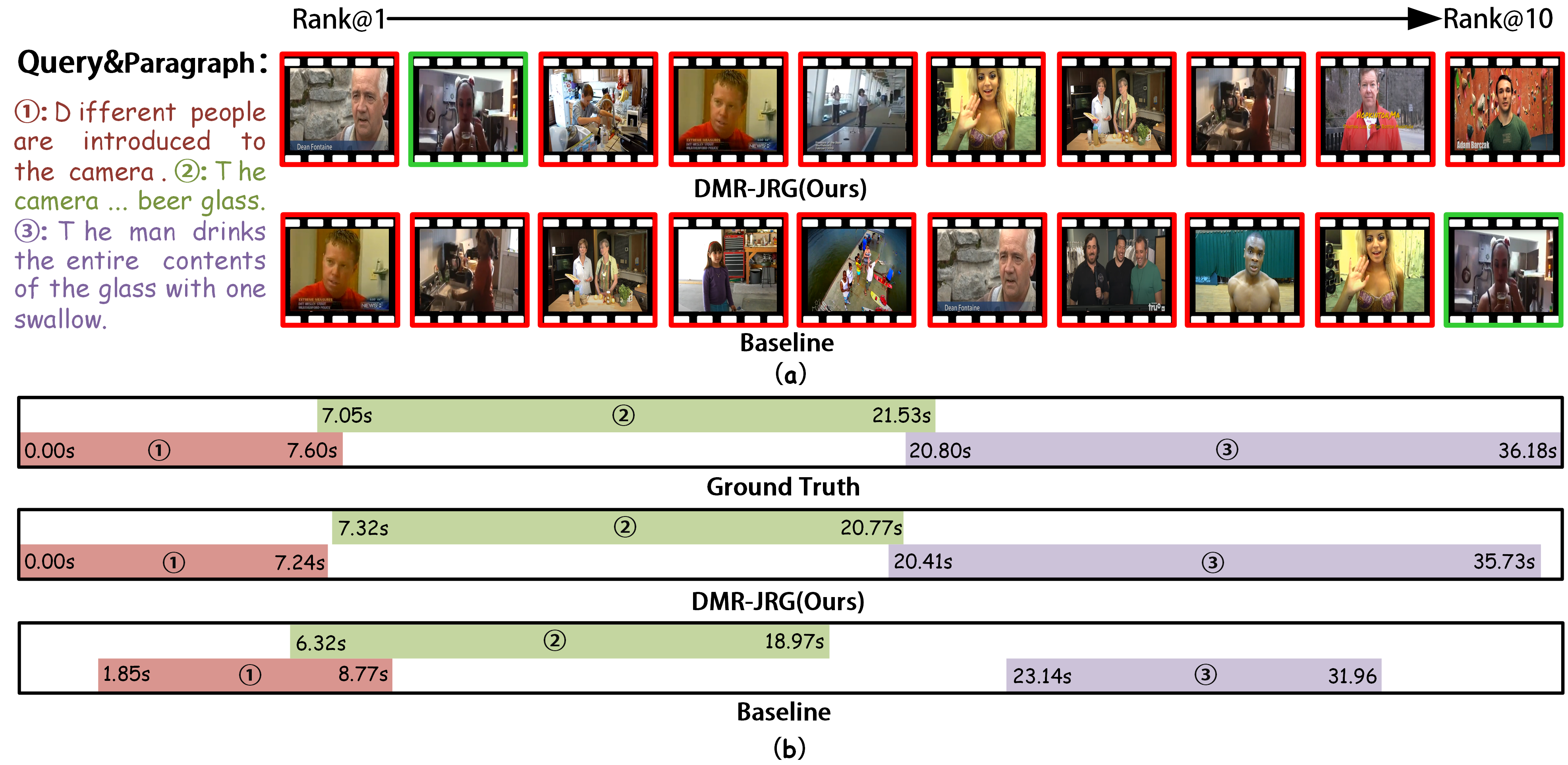}

	\caption{Visualization of prediction results from different models. In (a), the first row shows the top-10 retrieval results obtained by the proposed model, while the second row shows the top-10 retrieval results obtained by the baseline model. Each bounding box in the images represents a video. Green bounding boxes indicate correct retrieval results, while red bounding boxes indicate incorrect retrieval results.(b) displays the grounding results of different models based on the correctly retrieved videos, where Ground Truth represents the annotated temporal labels.} 
 
	\label{Fig09}
\end{figure*}

Furthermore, as shown in Table ~\ref{Table3}, the model performance improves progressively with the incremental incorporation of the VTC-LD, BTSE-TD, and VTFA-GD components in the grounding branch. Notably, in comparison to the ``Base+VPCMR'' model, the ``Base+VPCMR+VTC-LD+BTSE-TD+(VTFA-GD w/ GRRM)'' model, which incorporates multi-dimensional consistency across local, temporal, and global dimensions in the grounding branch, , exhibits enhanced performance. Specifically, it achieves improvements of 3.03\%, 12.27\%, 5.09\%, 12.17\%, 4.93\%, 9.39\% across all evaluation metrics, respectively. These results demonstrate that by combining the strengths of these three components, we can effectively mitigate the differences between visual and textual modalities, thereby improving the accuracy of VPRG.

\begin{table*}[h]  
	\centering  
	\caption{Performance comparison of the proposed method using ``Word2Vec+Bi-LSTM'' versus BERT as the text encoder on the ActivityNet Captions and Charades-STA datasets. Evaluation results for R@10 IoU=0.3, R@100 IoU=0.3, R@10 IoU=0.5, R@100 IoU=0.5, R@10 IoU=0.7, and R@100 IoU=0.7 are listed. Best and runner-up values are in bold and underlined, respectively.}  
	\label{Table6}  
	\small  
	\setlength{\tabcolsep}{2pt} 
	\renewcommand{\arraystretch}{1.4}  
	\begin{tabular}{c|c|cc|cc|cc||cc|cc|cc}  
		\hline  
		\hline  
		\multirow{3}{*}{Method} & \multirow{3}{*}{Text Encoder} & \multicolumn{6}{c||}{ActivityNet Caption} & \multicolumn{6}{c}{Charades-STA} \\  
		\cline{3-14}  
		& & \multicolumn{2}{c|}{IoU=0.3} & \multicolumn{2}{c|}{IoU=0.5} & \multicolumn{2}{c||}{IoU=0.7} & \multicolumn{2}{c|}{IoU=0.3} & \multicolumn{2}{c|}{IoU=0.5} & \multicolumn{2}{c}{IoU=0.7} \\  
		& & R@10 & R@100 & R@10 & R@100 & R@10 & R@100 & R@10 & R@100 & R@10 & R@100 & R@10 & R@100 \\  
		\hline  
		\textbf{DMR-JRG(Ours)} & BERT & \underline{15.23} & \underline{53.11} & \underline{13.86} & \underline{47.42} & \underline{8.29} & \underline{29.04} & \underline{10.27} & \underline{40.59} & \underline{9.31} & \underline{28.74} & \underline{6.25} & \underline{19.07} \\  
		\hline  
		\textbf{DMR-JRG(Ours)} & Word2Vec+Bi-LSTM & \textbf{16.19} & \textbf{55.54} & \textbf{14.05} & \textbf{48.04} & \textbf{8.65} & \textbf{29.22} & \textbf{11.39} & \textbf{42.07} & \textbf{10.24} & \textbf{30.49} & \textbf{6.72} & \textbf{19.64} \\  
		\hline  
		\hline  
	\end{tabular}  
\end{table*}

\subsection{Parameter Selection and Analysis}
In this section, we analyze the influence of the hyperparameters $Q$ and $\left\{w_i\right\}_{i=1}^{Q=3}$ in Eqs. (9) and (14) on model performance. Additionally, we conduct a hyperparameter analysis for the $\lambda$, $\Delta$ and $\beta_1$ in the InfoNCE loss, triple ranking loss, and cross-modal retrieval loss of the retrieval branch. When analyzing the effect of each hyperparameter, all others are held constant. All hyperparameter evaluations are conducted on the ActiveNet Captions dataset.

\textbf{Influence of $Q$.}  $Q$ represents the number of candidate moments. Fig. \ref{Fig07} illustrates the impact of different $Q$ values on model performance, with all evaluation metrics achieving their best values when $Q=3$. This result suggests that using fewer candidate moments fails to capture the key information in the video corresponding to the textual query. Although incorporating more candidate moments provides broader temporal information, it also introduces redundancy, weakening the discriminative capabilities of the fused feature $\bar{\bm{f}}_m$. Therefore, we set $Q = 3$ in the experiments.

\begin{figure}[t!]
	\centering
	\includegraphics[width=0.90\linewidth]{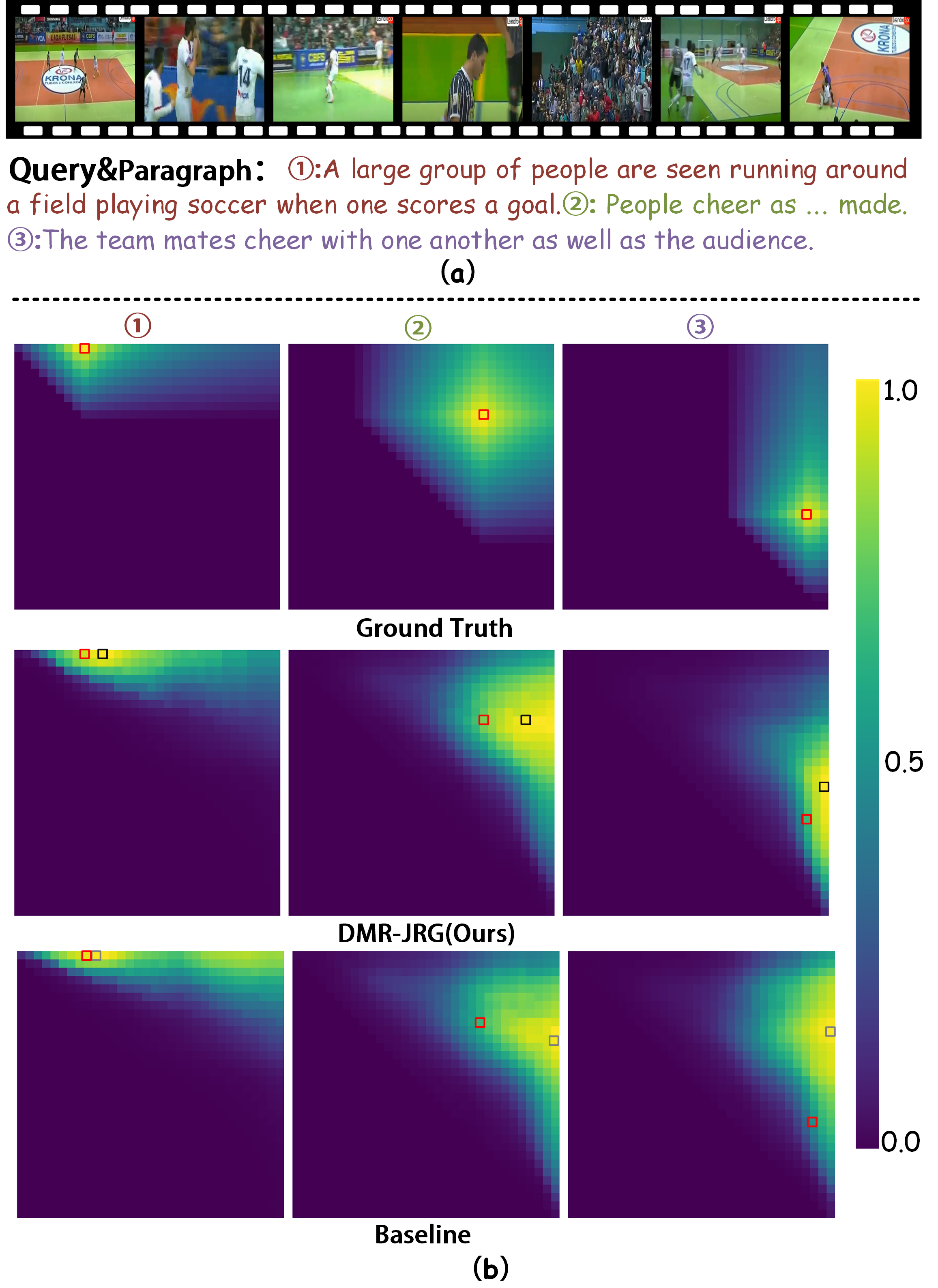}
	\caption{Visualization of the grounding score maps from different models. (a) shows the given paragraph query and its corresponding video. (b) shows the score maps $\{ {{\bm{P}}_m}\} _{m = 1}^M$ on ActivityNet Captions dataset. The first row is created from the ground truth data, the second row is the prediction results from our proposed model, and the third row is the prediction results from the baseline model. The brighter pixel areas indicate higher probability scores at the current proposal moment.}

	\label{Fig10}
\end{figure}

\textbf{Influence of $\left\{w_i\right\}_{i=1}^{Q=3}$.} $\left\{w_i\right\}_{i=1}^{Q=3}$ are balancing factors used to weight the features from different ranked candidate moments, aiming to obtain the visual features corresponding to the text query. This enables more precise cross-modal feature alignment on the global dimension (GD) by performing contrastive learning between videos and paragraphs. Table~\ref{Table4} shows the impact of different configurations of $\left\{w_i\right\}_{i=1}^{Q=3}$ on model performance. It is clearly observed that the model performs best when $\left\{w_i\right\}_{i=1}^{Q=3}$ are set to 0.4, 0.3, and 0.3. This configuration assigns a weight of 0.4 to the candidate moment ranked first, thereby highlighting the essential information. The remaining weights, each at 0.3, balance the information from other moments. This weighting strategy enhances model performance because the most confident candidate moment is given higher weight, ensuring the fused features are more accurate and discriminative. In contrast, when the weights are uniformly distributed as $\{0.33, 0.33, 0.33\}$, the model treats all candidate moments equally, ignoring their different rankings, which leads to suboptimal fusion and thus inaccurate visual features. Therefore, we set $\left\{w_i\right\}_{i=1}^{Q=3} = \{0.4, 0.3, 0.3\}$ in the experiments.

\textbf{Influence of $\lambda$, $\Delta$ and $\beta_1$.} 
Fig. \ref{Fig08} demonstrates the impact of varying values of $\lambda$, $\Delta$, and $\beta_1$ on model performance. $\lambda$ acts as a scaling factor in the InfoNCE loss, adjusting the similarity between positive and negative samples. Treating $\lambda$ as a learnable parameter allows the model to dynamically adapt its influence on each sample pair, leading to better alignment between video and text representations and improved retrieval performance. In contrast, fixing $\lambda$ hampers adaptability and leads to suboptimal performance. The parameter $\Delta$ in the triplet ranking loss defines the margin between positive and negative pairs. Setting $\Delta$ to 0.20 strikes an optimal balance: smaller margins weaken sample separation, while larger margins risk over-separation, increasing false negatives. $\beta_1$, which controls the trade-off between the InfoNCE and triplet ranking losses, was set to 0.04 based on previous work in JSG\cite{chen2023joint}. This value ensures a balanced contribution from both losses, facilitating effective model training. Higher values of $\beta_1$ overemphasize the InfoNCE loss, undermining the triplet ranking component, while lower values result in insufficient alignment between video and text features. In conclusion, the best performance is achieved when $\lambda$ is learned, $\Delta$ is set to 0.20, and $\beta_1$ is set to 0.04. These settings enable the model to effectively distinguish between positive and negative samples, thereby enhancing video-text retrieval.

\subsection{ Qualitative Results} Fig. \ref{Fig09} displays the retrieval and joint grounding results of our method compared to the baseline method on the ActivityNet Captions dataset. As shown in Fig. \ref{Fig09}(a), our proposed method achieves the correct match at the second rank, while the baseline model obtains the correct match at the tenth rank. This demonstrates the effectiveness of our method in the task of video paragraph retrieval. Additionally, the visualization of grounding results in Fig. \ref{Fig09}(b) indicates that our method's predicted paragraphs more closely align with the event timings of each sentence to the Ground Truth, whereas the baseline method shows deviations. This further confirms the effectiveness of our approach in video paragraph grounding.

\subsection{Further Discussion}

\textbf{Discussion of the grounding results.} Fig. \ref{Fig10} shows the grounding score maps $\{ {{\bm{P}}_m}\} _{m = 1}^M$ generated by our model compared to the baseline model on the ActivityNet Captions dataset. Fig. \ref{Fig10}(a) displays the given paragraph query and its corresponding video. In Fig. \ref{Fig10}(b), the first row displays the score maps generated for each sentence in the paragraph based on the Ground Truth. The second and third rows exhibit the score maps of the candidate moments predicted by our proposed DMR-JRG model and the Baseline model, respectively. For the $m$-th sentence in the paragraph, the brightest spot on its score map signifies the candidate moment with the highest matching score for this sentence. Within the Ground Truth score maps, red boxes denote the true positions of the candidate moments corresponding to each sentence. In the score maps predicted by our DMR-JRG method, black boxes highlight the best-matched  candidate moments, whereas gray boxes mark those predicted by the Baseline model. On the same score map, the distance between the black (or gray) box and the red box reflects the accuracy of the prediction: a closer distance indicates higher prediction accuracy, while a greater distance suggests lower accuracy.

From these results, we can observe that the DMR-JRG method predicts candidate moments that are closer to the ground truth moments, indicating superior grounding accuracy compared to the Baseline model. Additionally, further observation of the score maps corresponding to Ground Truth reveals that the events in the video clearly occur in a distinct temporal order, and the corresponding text descriptions are similarly ordered coherently. This time series characteristic is evident on the score maps. As the sentence number increases, the corresponding candidate moments in the video gradually shift further back in time. By observing Fig. \ref{Fig10}(b), we can clearly see the difference between the score maps predicted by the Baseline model and those predicted by the DMR-JRG model. The score maps predicted by the DMR-JRG model exhibit greater consistency with the time series characteristics of Ground Truth, further validating our method’s effectiveness in leveraging temporal dimension consistency to enhance performance.

\textbf{Discussion of the text encoder.} Following the SCN method \cite{lin2020weakly}, we construct a text encoder comprising Word2Vec \cite{mikolov2013distributed} for word embedding and Bi-LSTM \cite{hochreiter1997long} for text feature extraction. To evaluate the influence of different text encoders on model performance, we replace the “Word2Vec+Bi-LSTM” text encoder with BERT \cite{devlin2018bert} in the proposed network. As shown in Table~\ref{Table6}, our method demonstrates superior performance across all metrics when utilizing “Word2Vec+Bi-LSTM” as the text encoder, compared to using BERT as the text encoder. This superiority is attributed to the small scale of the dataset used in our weakly supervised VPRG task. Consequently, the ``Word2Vec+Bi-LSTM'' text encoder, with its simpler structure and fewer parameters, is easier to adjust and optimize, especially in weakly supervised learning scenarios with insufficient labeled data. In contrast, BERT, being a larger pre-trained model with a complex structure and numerous parameters, faces challenges in achieving optimal performance when trained on small-scale data samples. Furthermore, the limited data makes the model using BERT as text encoder prone to overfitting, thereby adversely affecting its generalizability.

\section{CONCLUSION}
In this study, we define a new task called VPRG. Specifically, the goal of VPRG is to retrieve the relevant video from a video corpus through a textual paragraph query and to locate the exact moments of each sentence in the paragraph within the video. To address the VPRG task, we propose a Dual-task Mutual Reinforcing Embedded Joint Video Paragraph Retrieval and Grounding method, termed DMR-JRG. We enhance the retrieval and grounding tasks mutually, rather than treating them as independent problems. This method mainly is divided into two branches: a retrieval branch and a grounding branch. The retrieval branch utilizes inter-video contrastive learning to coarsely align the global features of paragraphs and videos, constructing a coarse-grained feature space that facilitates the grounding branch in mining fine-grained contextual representations. The grounding branch explores the local, global, and temporal consistency between video segments and text paragraphs to build a fine-grained feature space. Additionally, we designed a grounding reinforcement retrieval module that brings the coarse-grained feature space of the retrieval branch closer to the fine-grained feature space of the grounding branch, thereby enhancing retrieval through improved grounding. Furthermore, the experiments on three challenging datasets have demonstrated the superior performance of our method.

Although the proposed method demonstrates distinct advantages, its performance may degrade substantially in unseen or complex scenarios. Future research will focus on exploring domain generalization techniques for VPRG to enhance the model's generalization capability in new scenarios. Additionally, we will investigate methods for modeling inter-object relationships to improve the accuracy of VPRG in challenging scenes involving multiple similar or overlapping objects.

\bibliography{mybibfile}
\end{document}